\documentclass[conference]{IEEEtran}
\IEEEoverridecommandlockouts
\usepackage{cite}
\usepackage{amsmath,amssymb,amsfonts}
\usepackage{algorithmic}
\usepackage{graphicx}
\usepackage{textcomp}
\usepackage{xcolor}
\usepackage{algorithm}
\usepackage{algorithmic}
\usepackage{subfigure}
\usepackage{booktabs}
\usepackage{marvosym}
\usepackage{multirow}

\def\BibTeX{{\rm B\kern-.05em{\sc i\kern-.025em b}\kern-.08em
    T\kern-.1667em\lower.7ex\hbox{E}\kern-.125emX}}
\begin{document}

\title{Deep Outdated Fact Detection in Knowledge Graphs\\

}

\author{\IEEEauthorblockN{Huiling Tu\IEEEauthorrefmark{1},
		Shuo Yu\IEEEauthorrefmark{2}\textsuperscript{(\Letter)}, Vidya Saikrishna\IEEEauthorrefmark{3},
		Feng Xia\IEEEauthorrefmark{4}\textsuperscript{(\Letter)}, and Karin Verspoor\IEEEauthorrefmark{4}}
	\IEEEauthorblockA{ \IEEEauthorrefmark{1}School of Software, Dalian University of Technology, Dalian 116620, China\\
 \IEEEauthorrefmark{2}School of Computer Science and Technology, Dalian University of Technology, Dalian 116024, China\\
 \IEEEauthorrefmark{3}Global Professional School, Federation University Australia, Ballarat, VIC 3353, Australia\\
		\IEEEauthorrefmark{4}School of Computing Technologies, RMIT University, Melbourne, VIC 3000, Australia\\
		{nataliathree@outlook.com}, {shuo.yu@ieee.org}, {v.saikrishna@federation.edu.au}, {f.xia@ieee.org}, {karin.verspoor@rmit.edu.au}
	}
}



\maketitle

\begin{abstract}
Knowledge graphs (KGs) have garnered significant attention for their vast potential across diverse domains. However, the issue of outdated facts poses a challenge to KGs, affecting their overall quality as real-world information evolves. Existing solutions for outdated fact detection often rely on manual recognition. In response, this paper presents DEAN (Deep outdatEd fAct detectioN), a novel deep learning-based framework designed to identify outdated facts within KGs. DEAN distinguishes itself by capturing implicit structural information among facts through comprehensive modeling of both entities and relations. To effectively uncover latent out-of-date information, DEAN employs a contrastive approach based on a pre-defined Relations-to-Nodes (R2N) graph, weighted by the number of entities. Experimental results demonstrate the effectiveness and superiority of DEAN over state-of-the-art baseline methods.

\end{abstract}


\section{Introduction}
Knowledge graphs (KGs)~\cite{peng2023kgsurvey,ji2021survey}, as a graph-based data structure, can represent factual triples organized in the form of $\langle head, relation, tail \rangle$. Various KGs (such as Wordnet and Freebase) have preserved structural information of nodes and edges representing real-world entities and relationships, respectively. The representation of KGs can be beneficial to enhance the efficiency of downstream applications, including data cleaning, question answering, recommendation, etc.

KGs serve as a powerful tool for modeling complexity and dynamics of real-world data. The factual knowledge represented by KGs is meant to be dynamic and adaptable to real-world changes. Even though existing KGs contain numerous facts, they still suffer from issues such as incompleteness and inaccuracies compared to real-world facts~\cite{10031054}. One of the most common issues with fact change is that facts tend to be outdated. Therefore, it is crucial to detect outdated facts to ensure the quality of KGs.

\begin{figure}[htbp]
	\centering
	\includegraphics[width=1\linewidth]{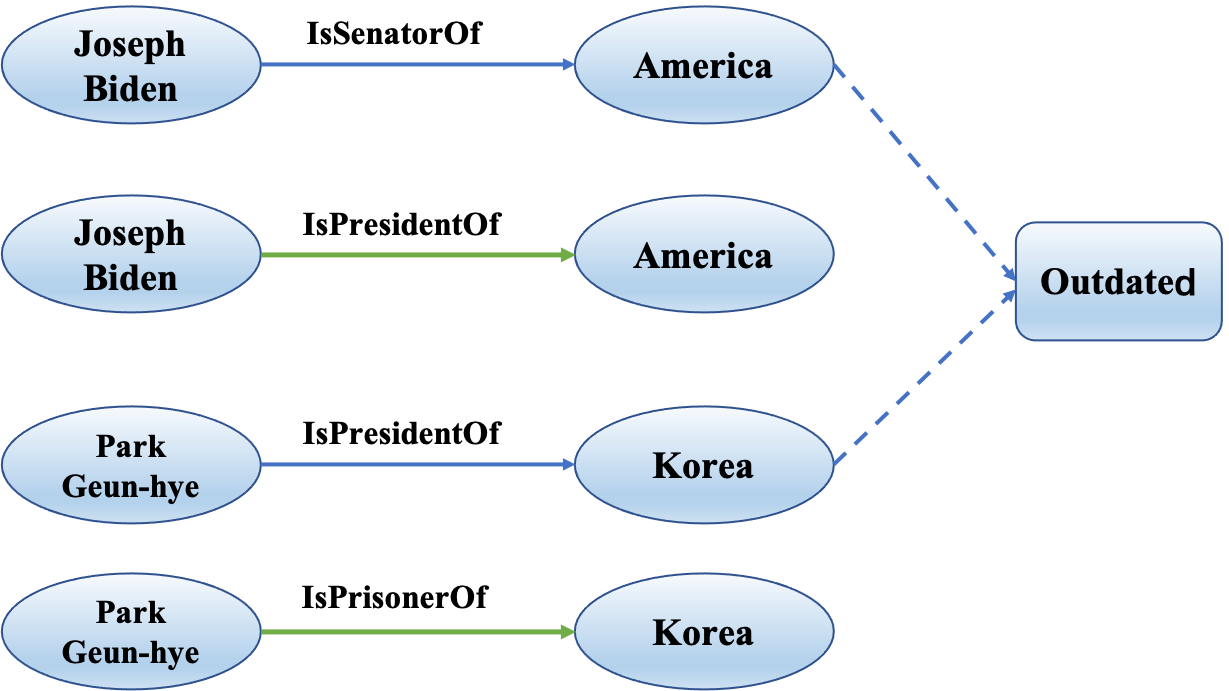}
	
	\caption{An example of outdated facts caused by relation changes}
	\label{fig1}
\end{figure}

Facts become outdated due to various reasons. Any changes in entities or relations can cause the triplets in the KGs to be outdated. For instance, as illustrated in Figure~\ref{fig1}, Joseph Biden is the \textbf{senator of} America. After the recent presidential election, Joseph Biden becomes the \textbf{president of} America. Obviously understandable, this change in a relationship turns the information that Joseph Biden is the senator of America, is outdated. Despite frequent updates to KGs, the existence of outdated facts is widespread.

Tran et al.~\cite{tran2013automatic} attempt to extract up-to-date facts from websites or text, which serve as reference data to detect outdated facts. Hao et al.~\cite{hao2020outdated} take human-in-the-loop into consideration, and leverage human recognition to guarantee the effectiveness of the outdated fact detection process. Previous studies heavily rely on prior knowledge from human to detect outdated facts. Moreover, implicit structural information of facts are not fully considered. Knowledge graph embedding methods have been proven effective in capturing implicit structural information~\cite{liu2022mirror,PU2023104464}. However, conventional embedding approaches cannot be directly and effectively applicable for detecting outdated facts~\cite{nguyen2022node, liu2022selfkg}. The reason is that outdated facts are not exactly equivalent to incorrect or anomalous facts and which cannot be distinguished from non-outdated facts by similarity.

To handle the above challenges, this paper first gives the formal definition of outdated fact and describe the outdated fact detection task. Then, we propose a novel deep learning-based framework, namely \textbf{DEAN} (\textbf{D}eep Outdat\textbf{E}d F\textbf{A}ct Detectio\textbf{N}), to detect outdated facts in the knowledge graph. DEAN first learns the representation of entities and relationships to preserve implicit structural information of knowledge through fact attention module~\cite{DBLP:conf/iclr/VelickovicCCRLB18}. In addition, to better differentiate between outdated facts and their corresponding non-outdated facts, we design a contrastive approach based on a pre-defined weighted graph, namely R2N (Relations-to-Nodes). As a result, DEAN is capable of extracting more expressive features for relations in the KGs. To authenticate the effectiveness of DEAN, we compare DEAN’s performance with the state-of-the-art approaches for knowledge graph embedding on the outdated fact detection task. The main contributions are summarized below:
\begin{itemize}
	\item \textbf{Problem fomulation:} We formally define the challenge of outdated facts arising from changes in relations within knowledge graphs, establishing the foundation for the outdated fact detection task.
	\item \textbf{Deep outdated fact detection - an innovative solution:} We propose a novel deep learning-based framework, namely DEAN, for addressing the outdated fact detection task. DEAN excels in extracting implicit structural information from facts, providing robust outcomes. The framework incorporates a novel contrastive approach utilizing a pre-defined weighted R2N graph.
	\item \textbf{Empirical validation:} We conduct extensive experiments across six datasets to validate the efficacy of DEAN. The results unequivocally showcase DEAN's superior performance compared to a spectrum of state-of-the-art baseline methods, establishing its prowess in outdated fact detection.
	
\end{itemize}  

\section{Related Work}
\label{rw}
In this section, we review the related work about outdated fact and knowledge graph embedding (KGE).
\subsection{Outdated Fact Detection}
Facts change dynamically as the real world constantly evolves. Therefore, discovering outdated facts and keeping them up-to-date is crucial for knowledge graph quality control. Furthermore, the updated facts would be more beneficial for many knowledge-driven applications. The study of outdated facts has already gained momentum. Tran et al.~\cite{tran2013automatic} extract the facts in the general web utilizing pattern learning. This work, aiming to address the outdated attribute values problem, gives a different definition of outdated facts. Similar to the research~\cite{tran2013automatic}, Tran et al.~\cite{6719874} propose an approach that is the extension of the pattern-based technique for detecting outdated attribute values, and it has more flexible patterns and utilizes  entity search for extracting related entities from websites. Liang et al.~\cite{DBLP:conf/ijcai/LiangZX17} settle the issue of keeping the knowledge base in sync with the online encyclopedia by designing a set of synchronization principles upon their developed system with a predictor. Hao et al.~\cite{hao2020outdated} propose an iterative outdated fact classification algorithm to explore the likelihood of each fact being outdated and develop a human-in-the-loop method utilizing the logical rule to detect the outdated facts in the knowledge base. Tang et al.~\cite{tang2019learning} take the implicit structural information of news text into consideration and develop a novel graph-based neural network approach incorporating the attention mechanism to keep the knowledge graph up-to-date. 

While the literature mentioned above only explicitly mines outdated facts, they overlook the implicit structural information within the facts. Therefore, this paper intends to capture and leverage the implicit structural information to improve the detection of outdated facts.
\subsection{Knowledge Graph Embedding}
Knowledge representation~\cite{DBLP:conf/kdd/HuangSDLL021} is the foundation of the knowledge-aware application. Recent years have observed expeditious growth in knowledge graph embedding. Existing knowledge graph embedding approaches can be roughly categorized into several groups: \emph{Translation-based methods}, \emph{Neural Networks-based methods}, \emph{Graph-based methods}~\cite{Xia2021TAI}, etc. TransE~\cite{DBLP:conf/nips/BordesUGWY13}, TransH~\cite{wang2014knowledge}, and TransR~\cite{DBLP:conf/aaai/LinLSLZ15} are the most characteristic translational models, all of them learn representations by embedding relations as translations from head to tail entities. In addition, neural network-based and graph-based approaches have also yielded significant performance in KGE research. For example, ConvE~\cite{DBLP:conf/aaai/DettmersMS018} applies 2D convolution on representation to complete missing facts in KGs and develops a scoring procedure to speed up the training process. Moreover, RGCN~\cite{DBLP:conf/esws/SchlichtkrullKB18} is an attempt at multi-relational graph scenarios, and it first utilizes the Graph Convolutional Network (GCN)~\cite{DBLP:conf/iclr/KipfW17} to deal with interactions between multiple relationships. Shang et al.~\cite{shang2019end} propose an end-to-end graph structure-sensitive convolutional network, the encoder of which utilizes weighted GCN to encode the attributes of graph nodes and capture the knowledge graph structure as input. Its decoder can preserve translation properties between entity-to-relational embeddings. 

Although the research mentioned above demonstrates strong performance in predicting missing facts, they cannot be directly applied to deal with outdated facts. Hence, this opens up an opportunity for us to propose a novel method that can automatically identify outdated facts in the KGs by taking the features of entities and relations into consideration.       

\section{Preliminaries}
\label{pre}
Before introducing our framework, we provide definitions utilized throughout this work, including knowledge graph, fact, and outdated fact. 
\subsection{Definitions}
\textbf{Knowledge graph.} We consider a knowledge graph as a multi-relational graph defined as $\mathcal{KG}=\{\mathcal{E}, \mathcal{R}, \mathcal{F}\}$, where $\mathcal{E} = \left\{e_{1}, e_{2}, \dots, e_{\left| \mathcal{E} \right|}\right\}$ denotes an entity type set containing $\left| \mathcal{E} \right|$ kinds of entities, $\mathcal{R} = \left\{r_{1}, r_{2}, \dots, r_{\left| \mathcal{R} \right|}\right\}$ represents the types set of relations, and $\mathcal{F} = \left\{f_{1}, f_{2}, \dots, f_{\left| \mathcal{F} \right|}\right\}$ denotes the set of fact triples in the KGs containing $\left| \mathcal{F} \right|$ facts.

\textbf{Fact.} In the KGs, a fact is a triple composed of a head entity, a tail entity, and a relation. We utilize lowercase letters $e_{h}$, $r$, and $e_{t}$ to act in place of the head entity, relation, and tail entity, respectively. Note that, $r$ can be a symbolic representation of multiple relation types. Hence, we consider that a fact is described symbolically as $\langle e_{h}, r, e_{t}\rangle \in \mathcal{F} $. An example of a fact can be $<$London, IsCapitalOf, England$>$. 

\textbf{Outdated fact.} We consider that a fact $f_{o}=\langle e_{h}, r_o, e_{t}\rangle$ is defined as an \emph{outdated fact} with respect to other updated facts $f'=\langle e_{h}, r', e_{t}\rangle$ on the same head and tail entities  ($e_{h}, e_{t}$) but $r_o \neq r'$. Due to real-world information delays, we think that the frequency of knowledge graph construction or update is not guaranteed to make the data always real-time and completely accurate. Hence, in this work, we consider facts being outdated statically rather than dynamically. In this way, we could believe that the outdated fact $f_{o}$ and its corresponding non-outdated facts can co-exist in $\mathcal{KG}$ at the same time.

\subsection{Task Description}
In this work, the outdated fact detection problem is formulated as a binary classification problem. Specifically, the fact is either outdated or not for each fact in $\mathcal{KG}$. Therefore, we annotate each fact in the datasets we utilize. For example, if a fact is an outdated fact, we label the fact as 0. Otherwise, we label it as 1. These annotations are utilized for model training and detection.

Based on the aforementioned definitions, the outdated fact detection task aims to determine whether a fact in $\mathcal{KG}$ is outdated. The input for this task consists of all the training facts, which need to be modeled to extract effective feature information about entities and relations. The information is then utilized to develop an effective classification approach that can accurately identify whether a fact is outdated or not. For the output of this task, we attempt to obtain the predicted labels for triples through the learning and training process. That is, if given a partial fact and its true labels (0 or 1), DEAN is capable of detecting whether the remaining facts are out-of-date.

\tablename{~\ref{tab1}}  lists all the detailed descriptions of symbols utilized in this work.
\begin{table}[htbp]
	\centering
	\caption{Descriptions of symbols}
	\begin{tabular}{ll}
		\toprule
		Notations & Descriptions\\
		\midrule 
		$\mathcal{KG}$  & The knowledge graph  \\
		$\mathcal{E}$  &  The entity set \\
		$\mathcal{R}$  & The relation set\\
		$\mathcal{F}$  & The set of facts \\
		$n_{1}=\left| \mathcal{E} \right|$& The number of entity types \\
		$n_{2}=\left| \mathcal{R} \right|$& The number of relation types \\
		$n_{3}=\left| \mathcal{F} \right|$& The total number of facts in the KGs \\
		$d_{1}$&The dimension of entity features\\
		$d_{2}$&The dimension of relation features\\
		$\langle e_{h}, r, e_{t}\rangle$ & A fact of head entity, relation, tail entity\\
		$f_{o}$& An outdated fact\\
		$\boldsymbol{E} \in \mathbb{R}^{n_{1} \times d_{1}}$&The entity feature matrix \\
		$\boldsymbol{R}\in \mathbb{R}^{n_{2} \times d_{2}}$&The relation feature matrix \\
		$\boldsymbol{A}$& The adjacency matrix of R2N graph\\
		$\boldsymbol{X}$ & The node feature matrix of R2N graph\\
		$G_{R2N}$&  The R2N graph\\
		$E$ & The set of edges in the R2N graph\\
		$K$ & The number of head in attention mechanism\\
		$\lambda$ & The coefficient of the loss function\\
		\bottomrule
	\end{tabular}
	\label{tab1}
\end{table}

\section{Design of DEAN}
\label{framework}
\begin{figure*}[htbp]
	\centering
	\includegraphics[width=1\linewidth]{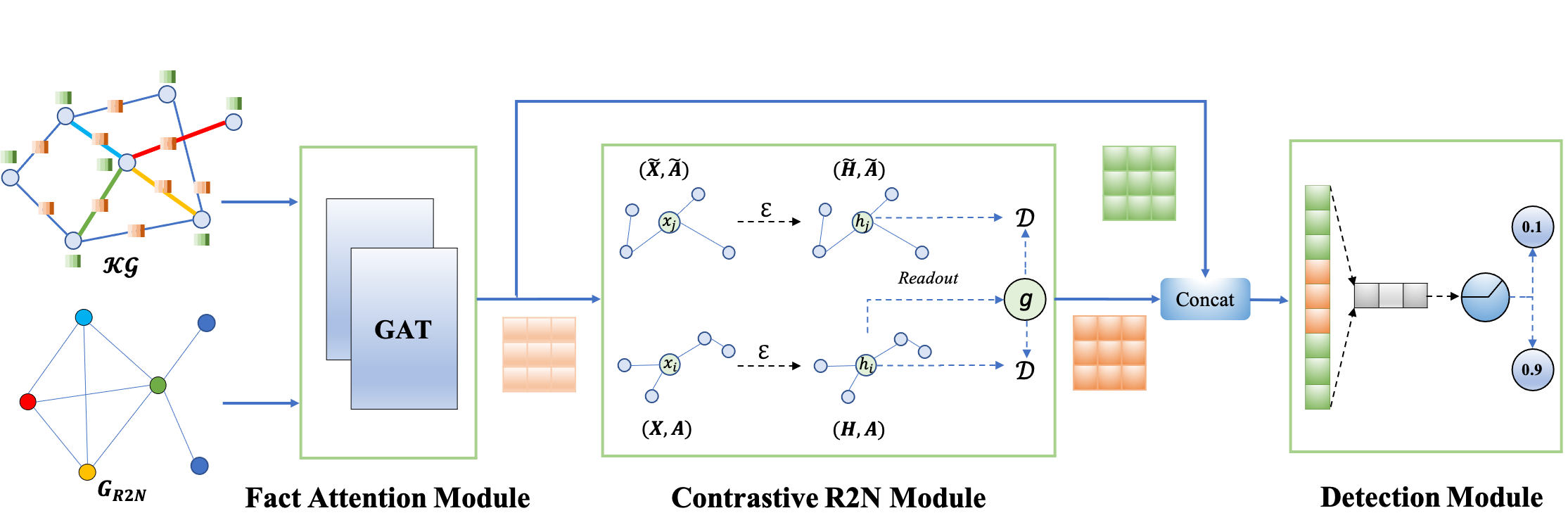}
	
	\caption{The overall framework of DEAN.}
	\label{fig2}
\end{figure*}

In this section, we describe the framework of the proposed approach DEAN for outdated fact detection in detail, as illustrated in Figure~\ref{fig2}. Specifically, DEAN is composed of three main components, namely \emph{Fact Attention Module}, \emph{Contrastive R2N Module}, and \emph{Detection Module}. Firstly, the knowledge graph embedding technique can effectively learn the complex structural information of multi-relational data. Hence, this paper adopts this technique to address the outdated fact problem. Specifically, we utilize the graph attention mechanism for preserving the knowledge graph's structural information in the fact attention module. Second, since an outdated fact due to relationship changes is highly similar to its corresponding non-outdated fact, it would lead to a great challenge for the detection process. In order to handle the challenge, we design a contrastive R2N module that can better distinguish between information differences in relationships and strengthen the learning feature for each relation in the KGs. This module requires a specific weighted graph (R2N) to be pre-generated based on the knowledge graph, and its construction algorithm is shown in Algorithm~\ref{al1}. In this way, we can incorporate embeddings for both entities and relations. Afterwards, we can obtain the fact embeddings. Finally, \emph{Detection Module} detects outdated facts. We train a binary classification model with FCN to acquire the predicted labels of all facts, allowing for more automated outdated fact detection. We then provide detailed descriptions of \emph{Fact Attention Module} and \emph{Contrastive R2N Module}, followed by the training objective and an analysis of the time complexity of our framework. 
  
  \begin{algorithm}[h]
  	\centering
  	\caption{R2N Graph Construction}
  	\label{al1}
  	\begin{algorithmic}[1]
  		\REQUIRE A knowledge graph $\mathcal{KG}$
  		\ENSURE The adjacency matrix $\boldsymbol{A}$ of R2N graph
  		\STATE Generate a sparse matrix of entities $DA_{\mathcal{KG}}$ from triples $\mathcal{F}=\{\langle e_{h}, r, e_{t}\rangle\}$;
  		\STATE Initialize the adjacency matrix $\boldsymbol{A} \in \mathbb{R}^{n_{2}\times n_{2}}$ of R2N graph  filled with 0;
  		\FOR{$e_i \in DA_{\mathcal{KG}}.keys$}
  		\STATE $R_i \leftarrow DA_{\mathcal{KG}}[e_i]$; 
  		\FOR{$(r_x, r_z) \in R_i$}
  		\STATE $\boldsymbol{A}[x, z] += 1$; 
  		\STATE $\boldsymbol{A}[z, x] += 1$;
  		\ENDFOR
  		\ENDFOR
  		\RETURN $\boldsymbol{A}$ 
  	\end{algorithmic}
  \end{algorithm}
  
\subsection{Fact Attention Module}
Inspired by~\cite{DBLP:conf/iclr/VelickovicCCRLB18, DBLP:conf/acl/NathaniCSK19, liu2022selfkg, yu2023spatio, xia2022cengcn}, we adopt the attention mechanism for feature learning in this work. Before we begin to start the feature learning process, for both entity and relation types in a $\mathcal{KG}=\{\mathcal{E}, \mathcal{R}, \mathcal{F}\}$, we train the TransE~\cite{DBLP:conf/nips/BordesUGWY13} to generate the initial features for both entities and relations as the input for fact attention module. Therefore, the initial entity features are represented by $\boldsymbol{E}=\left\{\boldsymbol{e}_{1}, \boldsymbol{e}_{2} \ldots, \boldsymbol{e}_{n_{1}}\right\}\in \mathbb{R}^{n_{1} \times d_{1}} $, where $\boldsymbol{e}_{i}$ is the feature vector for entity $e_{i}$, $n_{1}=\left| \mathcal{E} \right|$ denotes the number of entity types and $d_{1}$ represents the dimension of each entity feature vector. The initial relation features $\boldsymbol{R}=\left\{\boldsymbol{r}_{1}, \boldsymbol{r}_{2} \ldots, \boldsymbol{r}_{n_{2}}\right\}\in \mathbb{R}^{n_{2} \times d_{2}}$ is also generated here by training TransE, where $\boldsymbol{r}_{x}$ is the feature vector for relation $r_{x}$, $n_{2}=\left| \mathcal{R} \right|$ denotes the number of relation types and each relation is mapped to $d_{2}$ dimensions. 

\begin{figure}[htbp]
	\centering
	\includegraphics[width=1\linewidth]{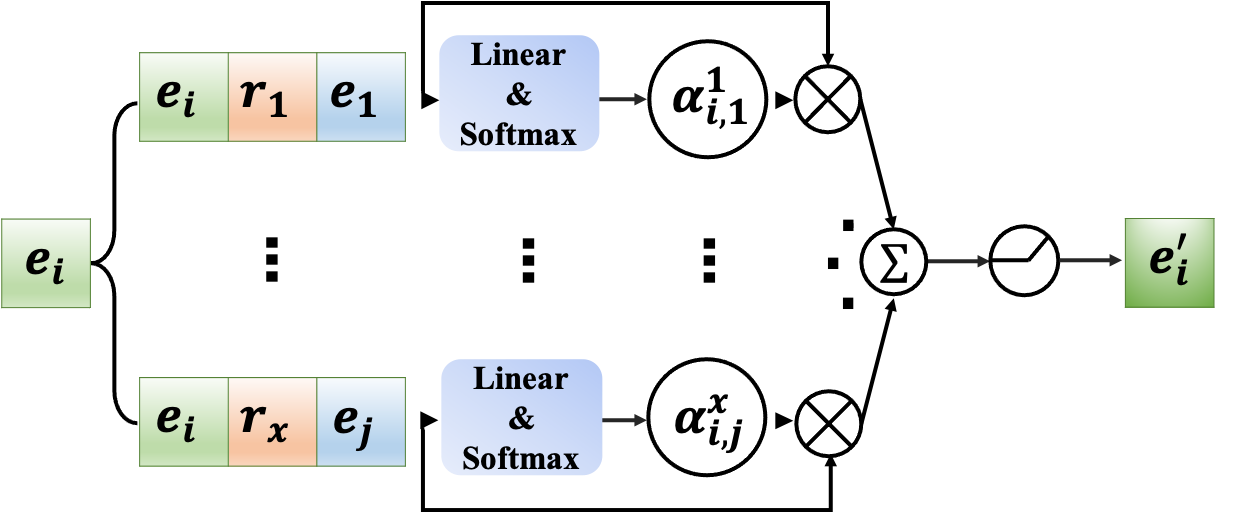}
	
	\caption{The computation of attention mechanism.}
	\label{fig3}
\end{figure}

In order to acquire more expressive feature information for entities and relations from the input knowledge graph, we build the fact attention module as part of the feature learning process. Specifically, to obtain the updated low-dimensional latent representation of entity $e_{i}$, we first need to learn the feature of each fact related to $e_{i}$. The representation of each fact connected with $e_{i}$ is performed by a linear transformation over the concatenation of entity and relation feature vectors. As an example of fact $f_{i j}^{(x)}=\left(e_{i}, r_{x}, e_{j}\right)$, we can get its representation according to the following equation:\begin{equation}
	\boldsymbol{f}_{i j}^{(x)}=\boldsymbol{\Theta}_{1}\left[\boldsymbol{e}_{i}\left\|\boldsymbol{r}_{x}\right\| \boldsymbol{e}_{j}\right],
\end{equation}
where $\boldsymbol{f}_{i j}^{(x)}$ represents the representation of fact $f_{i j}^{(x)}$, and $\boldsymbol{\Theta}_{1}$ as a learnable parametric matrix. Then, one learnable shared weight matrix is required to compute the attention scores for each fact contained $e_{i}$. To that end, a matrix $\boldsymbol{\Theta}_{2}$ is applied to every fact. Meanwhile, the softmax function is used to normalize each transformed fact representation. Figure~\ref{fig3} shows the calculation process of attention score. Hence, taking the fact $f_{i j}^{(x)}$as an example, the attention score can be calculated as follows:
\begin{equation}
	\begin{aligned}
		a_{i j}^{(x)} &= \operatornamewithlimits{softmax}_{j,x} \left(\sigma\left(\boldsymbol{\Theta}_{2} \boldsymbol{f}_{i j}^{(x)}\right)\right)\\
		&=\frac{\exp \left(\sigma\left(\boldsymbol{\Theta}_{2} \boldsymbol{f}_{i j}^{(x)}\right)\right)}{\sum\limits_{n \in N_{i}} \sum \limits_{z \in \mathcal{R}_{i n}} \exp \left(\sigma\left(\boldsymbol{\Theta}_{2} \boldsymbol{f}_{i n}^{(z)}\right)\right)},
	\end{aligned}
\end{equation}
where $N_{i}$ is the set of neighbours of entity, $e_{i}$, ${R}_{i n}$ represents the relation set between entities $e_{i}$ and $e_{n}$, and $\sigma$ denotes parametric LeakyReLU function. Once the attention scores of all facts associated with $e_{i}$ are obtained, they are utilized for updating feature information of $e_{i}$. The new feature vector of entity $e_{i}$ is the sum of each fact embedding weighted by its corresponding attention score expressed as follows:
\begin{equation}
	\boldsymbol{e}_{i}^{\prime}=\sigma\left(\sum_{j \in \mathcal{N}_{i}} \sum_{x \in \mathcal{R}_{i j}} a_{i j}^{(x)} \boldsymbol{f}_{i j}^{(x)}\right).
\end{equation}
The \emph{multi-head attention}~\cite{DBLP:conf/iclr/VelickovicCCRLB18} is also introduced to stabilize the learning process. The multi-head attention with K heads performs the following equation, resulting in the representation:
\begin{equation}
	\boldsymbol{e}_{i}^{\prime}=\mathop{\Vert}\limits_{k=1}^{K} \sigma\left(\sum_{j \in \mathcal{N}_{i}} a_{i j}^{(x), k}  \boldsymbol{f}_{i j}^{(x), k}\right),
\end{equation}
where $\mathop{\Vert}$ denotes the concatenation operation. Especially in the final layer of this module, the average operation is employed to acquire the final entity embeddings expressed as follows:
\begin{equation}
	\boldsymbol{e}_{i}^{\prime}=\sigma\left(\frac{1}{K} \sum_{k=1}^{K} \sum_{j \in \mathcal{N}_{i}}\sum_{x \in \mathcal{R}_{i j}} a_{i j}^{(x), k} \boldsymbol f_{i j}^{(x), k}\right),
\end{equation}
where $\boldsymbol e_{i}^{\prime} \in \mathbb{R}^{d_{1}^{\prime}}$ and $d_{1}^{\prime}$ represents the final embedding dimension of entities.
And we define the loss function as:
\begin{equation}
	\label{lgat}
	\mathcal{L}_{gat}=\sum_{f_{ij} \in S} \sum_{f_{ij}^{\prime} \in S^{\prime}} max \left(d_{f_{i j}^{\prime}}-d_{f_{ij}}+\beta, 0 \right),
\end{equation}
where $S$ denotes the set of valid facts and $S^{\prime}$ represents the set of invalid facts based on $S$, $\beta>0$ is a hyperparameter to fine-tune the weight of this layer. Here, $d_{f_{ij}}=\left\|\boldsymbol{e}_{i}+\boldsymbol{r}_{x}- \boldsymbol{e}_{j}\right\|_1$ denotes embedding distance from entity ${e}_{i}$ to entity ${e}_{j}$ by utilizing L1-norm dissimilarity measure.

\subsection{Contrastive R2N Graph Module}
Since an outdated fact due to relations change is highly similar to its corresponding non-outdated fact, it would lead to a great challenge for the detection process. However, existing techniques cannot deal with this challenge well. Based on what was mentioned before and motivated by previous research~\cite{velickovic2019deep}, we design a Contrastive R2N module to strengthen the learning of relation representations for addressing the above-mentioned challenge. 

We first leverage the proposed Algorithm~\ref{al1} to obtain the R2N graph structural information for the contrastive R2N module. As shown in Algorithm~\ref{al1}, the nodes of the R2N graph are converted based on the relationship of the knowledge graph. Therefore, the node features of the R2N graph are also obtained based on the relations feature information of the knowledge graph. We perform a linear transformation on the relational information by utilizing a weight matrix expressed as follows:
\begin{equation}
	\boldsymbol{R}^{\prime}=\boldsymbol{R} \boldsymbol{\Theta}_{3},
\end{equation}
where $\boldsymbol{R}^{\prime} \in \mathbb{R}^{n_{2} \times d_{2}^{\prime}} $ and $d_{2}^{\prime}$ represents the  embedding dimension of relations. Thus, the R2N graph can be defined as a weighted undirected graph $G_{R2N} = (R, E)$, where $R$ is the node set of R2N graph based on relation types set in $\mathcal{KG}$, and $E$ is the edge set. $\boldsymbol{A} \in \mathbb{R}^{n_{2} \times n_{2}}$ represents the adjacency matrix of R2N graph. In the R2N graph, we can get the node feature set represented as $\boldsymbol{X}=\left\{\boldsymbol{x}_{1}, \boldsymbol{x}_{2} \ldots, \boldsymbol{x}_{n_{2}}\right\} $, $\boldsymbol{X}=\boldsymbol{R}^{\prime}$. Furthermore, we consider a multi-layer graph convolutional
network as the encoder, and the message propagation rule can be formulated as follows:
\begin{equation}
	\label{cccc}
	\boldsymbol{H}^{(l)}=\sigma \left( \boldsymbol{\hat{D}}^{-\frac{1}{2}} \boldsymbol{\hat{A}} \boldsymbol{\hat{D}}^{-\frac{1}{2}} \boldsymbol{H}^{(l-1)}\boldsymbol{\Theta}^{(l-1)}
	\right),
\end{equation}
where $\boldsymbol{\hat{A}}=\boldsymbol{A}+\boldsymbol{I}_{N}$ contains with inserted self-loop and $\boldsymbol{\hat{D}}$ denotes degree matrix of $\boldsymbol{\hat{A}}$. In addition, $\sigma$ is the PReLU function, $\boldsymbol{\Theta}^{(l-1)}$ is a learnable weight matrix in the $(l-1)$-th GCN layer, and $\boldsymbol{H}^{(l)}$ is the feature in the $l$-th GCN layer. Specially, $\boldsymbol{H}^{(0)}=\boldsymbol{X}$ and we denote $\boldsymbol{H}^{(L)}$ as the learning feature matrix of final GCN layer.

In order to capture more expressive relational representation, we utilize a contrastive method to boost the node representations in the R2N graph. To that end, we construct a negative R2N graph, where the adjacency matrix $\tilde{\boldsymbol{A}}$ is the same as $\boldsymbol{A}$ while node features $\tilde{\boldsymbol{X}}$ are obtained by randomly shuffling features $\boldsymbol{X}$. The information aggregation process is the same as Eq.(\ref{cccc}). Moreover, R2N graph-level representation is obtained through a readout function: $\mathbb{R}^{n_{2} \times d_{2}^{\prime}} \to \mathbb{R}^{d_{2}^{\prime}}$. Specifically, the function is written as follows:
\begin{equation}
	\boldsymbol{g}=Readout(\boldsymbol{H}^{(L)})=\sum_{t=1}^{n_{2}} \frac{\boldsymbol{h}_{t}^{(L)}}{n_{2}},
\end{equation}
where $\boldsymbol{h}_{t}^{(L)}$ is $t$-th row of $\boldsymbol{H}^{(L)}$ and $n_{2}$ is the number of positive samples. Thus, after obtaining positive, negative sample pairs and graph-level representation, a discriminator $\mathfrak{D}$ is employed to judge positive and negative labels, $\mathfrak{D}$: $\mathbb{R}^{d_{2}^{\prime}} \times \mathbb{R}^{d_{2}^{\prime}} \to \mathbb{R}$. It is formulated as follows:
\begin{equation}
	\mathfrak{D} \left(\boldsymbol{h}_{t}^{(L)} ,\boldsymbol{g}\right)=\sigma \left(\boldsymbol{h}_{t}^{(L)} \boldsymbol{\Theta^{d}}\boldsymbol{g}\right)
\end{equation}
where $\boldsymbol{\Theta^{d}}$ is a learnable parameter matrix of discriminator. In order to enhance representation, the objective function is designed to maximize the difference between positive and negative samples. Therefore, the objective function of this part utilizes a standard binary cross-entropy (BCE) loss between the  positive and negative samples, and is written as follows:
\begin{small}
	\begin{equation}
		\label{lr2n}
		\begin{split}
			\mathcal{L}_{R2N}=\frac{1}{\left(P+N\right)}(\sum_{i=1}^{P}\mathbb{E}_{(\boldsymbol{X},\boldsymbol{A})}[\log\mathfrak{D} \left(\boldsymbol{h}_{t}^{(L)},\boldsymbol{g}\right)] \\
			+\sum_{j=1}^{N} \mathbb{E}_{(\tilde{\boldsymbol{X}},\tilde{\boldsymbol{A}})}[1-\log\mathfrak{D} \left(\tilde{\boldsymbol{h}}_{t}^{(L)},\boldsymbol{g}\right)])
		\end{split},
	\end{equation} 
\end{small}where $P=n_{2}$ and $N$ represents the number of negative samples. 
\subsection{Training Objective}
In this subsection, we present the objective functions utilized in DEAN.

To enhance performance, we jointly train the fact attention module and contrastive R2N module, and the final loss function is calculated as follows:
\begin{equation}
	\mathcal{L}_{kge}=\mathcal{L}_{gat}+\lambda \mathcal{L}_{R2N},
\end{equation}
where $\lambda >0$ represents a hyperparameter.

Since we formulate the outdated fact detection task as a binary classification task, we utilize a simple classification model consisting of fully connected layers to detect outdated facts, and we also use the BCE loss for the detection process. The objective function is expressed as:
\begin{equation}
	\label{yyyyyy}
	\mathcal{L}=-\left[y\log \hat{y}+\left(1-y \right)\log \left(1-\hat{y} \right)\right],
\end{equation}
where $\hat{y}$ denotes the predicted label, and $y$ represents the true label. Specifically, in the experiment, if a fact is an outdated fact, we set $y=0$. Otherwise, $y=1$.

The overall framework of  DEAN is summarized
in Algorithm~\ref{al2}.

\begin{algorithm}[htb]
	\caption{Outdated Fact Detection}
	\label{al2}
	\begin{algorithmic}[1]
		\REQUIRE 
		$\mathcal{KG}$, $\boldsymbol{E}$, $\boldsymbol{R}$, Number of training epochs: $T_1, T_2$
		\ENSURE The set of predicted label for each fact in $\mathcal{KG}$: $\mathcal{\hat{Y}}$ 
		\STATE Obtain  adjacency matrix $\boldsymbol{A}$ of R2N graph utilizing Algorithm~\ref{al1};\STATE Divide the dataset into training sets and test sets, the set of true label $\mathcal{Y}$;\STATE // Feature learning\STATE Randomly initialize all weight parameters of knowledge graph representaion learning $\left\{\boldsymbol{\Theta}_{i}\right\}_{1\leq i \leq 3}$, $\boldsymbol{\Theta}^{d}$  and $\left\{\boldsymbol{\Theta}^{l}\right\}_{0\leq l \leq L}$;\\
		\FOR{$t_1 \in 1, 2, ..., T_1$}
		\STATE Caculate $\mathcal{L}_{gat}$ via Eq.(\ref{lgat});
		\STATE Caculate $\mathcal{L}_{R2N}$ via Eq.(\ref{lr2n});
		\STATE Caculate $\mathcal{L}_{kge}  \leftarrow \mathcal{L}_{gat}+\lambda \mathcal{L}_{R2N}$;
		\STATE Back propagation and update all parameters of this process;
		\ENDFOR
		\STATE // Outdated fact detection
		\FOR{$t_2 \in 1, 2, ..., T_2$}
		\STATE  Caculate $\mathcal{L}$ via Eq.(\ref{yyyyyy});
		\STATE Minimize $\mathcal{L}$ by Adam;
		\ENDFOR
		\RETURN $\mathcal{\hat{Y}}$ 
	\end{algorithmic}
\end{algorithm}
\subsection{Complexity Analysis}
We analyze the time complexity of DEAN by taking three main components into consideration respectively. For the fact attention module, the time complexity is $O(n_{1}d_{1}d_{1}^{\prime}+c_{a}d_{2}d_{2}^{\prime})$, where $c_{a}$ denotes the number of edges in the knowledge graph. In addition, the contrastive R2N module includes R2N graph construction and GCN information aggregation for relation learning. Hence, based on Algorithm~\ref{al1}, the time complexity of the R2N construction is $O(n_{2}{n_{1}}^2)$. The time complexity of GCN information aggregation is $O( c_{n})$, where $c_{n}$ denotes the number of edges in the R2N graph. Thus, the whole time complexity for this component is $O(n_{1}d_{1}d_{1}^{\prime}+c_{a}d_{2}d_{2}^{\prime}+n_{2}{n_{1}}^2+c_{n})$. The time complexity for the detection module is far less than the previously mentioned part, so we ignore this item. All in all, the overall time complexity of DEAN is $O(n_{2}{n_{1}}^2+n_{1}d_{1}d_{1}^{\prime}+c_{a}d_{2}d_{2}^{\prime}+c_{n})$.
\section{Experiments}
\label{experiment}
In this section, experiments are conducted to evaluate the performance of the DEAN framework. We first introduce the datasets utilized for experiments. Then, we describe the baselines and metrics. After that, we evaluate the effectiveness of the proposed approach by baseline algorithms and parameter sensitivity analysis.
\subsection{Datasets}
The existing datasets lack outdated facts. In this work, we try to simulate outdated facts as much as possible. Motivated by previous study~\cite{xie2018does, ren2022graph}, to better deal with the problem and make the performance evaluation of outdated facts detection more automatic, we perform experiments on known datasets and inject them with outdated facts. We select six datasets for the knowledge graph completion task from previous literature, and the details of the six datasets are described as follows:
\begin{itemize}
	\item  \emph{WN18}~\cite{DBLP:conf/nips/BordesUGWY13} is the subset of WordNet~\footnote{https://wordnet.princeton.edu}, which is composed of 40943 entities and 18 relations. WordNet is a lexical database that provides a definition for each synset and records the semantic relationships between different synsets.
	\item  \emph{WN18RR}~\cite{DBLP:conf/aaai/DettmersMS018} is generated from \emph{WN18}. The dataset \emph{WN18RR} contains 40943 entities and 11 relations, and it is aimed to deal with the inverse relation test leakage on the evaluation dataset of \emph{WN18}.
	\item \emph{FB15K-237}~\cite{toutanova2015observed} encompasses 14541 entities and 237 relations. The facts in this dataset are a subset of \emph{FB15K}  originated from  Freebase~\footnote{https://developers.google.com/freebase/}, and its inverse relations are removed.
	\item \emph{Kinship}, \emph{UMLS}, and \emph{Nations} are utilized the same as those preprocessed by ConvE~\cite{DBLP:conf/aaai/DettmersMS018}. \emph{Kinship} describes the kinship between characters, \emph{UMLS} describes connections between medical concepts.
\end{itemize}

Before conducting experiments, we performed a statistical analysis on the six datasets. For WN18RR and FB15K-237, we found few entities in its test and validation sets do not appear in the training set. Hence we performed a secondary cleaning on the above datasets and added the synthesized outdated facts into the training, test, and validation set of cleaned datasets, respectively. Furthermore, the outdated facts generated by us do not exist in the cleaned datasets. For each dataset, the number of outdated facts is about 20\% of the total data. According to Section~\ref{pre}, we generate outdated facts by adding a relation between the head and tail entity of the fact in the knowledge graph. For example, the set of all relation types existing between entities $h$ and $t$ is denoted as $\mathcal{R}_{(h,t)}$, the sets of relation types related to $h$ and $t$ are denoted as $\mathcal{R}_{h}$ and $\mathcal{R}_{t}$ respectively. Finally, the added relation between $h$ and $t$ we select is randomly from $\mathcal{R}^{*}$, where $\mathcal{R}^{*}=\mathcal{R}_{h} \cup \mathcal{R}_{t}-\mathcal{R}_{(h,t)}$. In the above manner, we can represent the concept of ``outdated'' more reasonably from the perspective of deep learning and semantics. \tablename{~\ref{tab2}} lists the statistics of the cleaned datasets.

\begin{table}[htbp]
	\centering
	\caption{Statistics of cleaned datasets}
	\setlength{\tabcolsep}{0.75mm}{
		\begin{tabular}{llllll}
			\toprule
			Dataset   & Entities & Relations & Training & Testing & Validation \\  \midrule
			WN18      & 40,943    & 18        & 141,442   & 5,000    & 5,000       \\
			WN18RR    & 40,559    & 11        & 86,835    & 2,924    & 2,824       \\
			FB15K-237 & 14,505    & 237       & 272,115   & 20,438   & 17,526      \\
			Kinship   & 104      & 25        & 8,544     & 1,074    & 1,068       \\
			UMLS      & 135      & 46        & 5,216     & 661     & 652        \\
			Nations   & 14       & 55        & 1,592     & 201     & 199        \\ \bottomrule
		\end{tabular}
		\label{tab2}
	}
\end{table}
\begin{table*}[t]
	\centering
	\caption{OUTDATED FACT DETECTION PERFORMANCE OF DIFFERENT APPROACHES. ``-'' DENOTES OUT OF MEMORY.}
	\resizebox{\textwidth}{!}{%
		\begin{tabular}{@{}cccccccccc@{}}
			\toprule
			Datasets                   & Metrics   & DisMult & ComplEx & RotatE & pRotatE & PairRE & CompGCN         & KBGAT              & \textbf{DEAN(ours)}     \\ \midrule
			\multirow{4}{*}{WN18}      & Accuracy  & 0.6752  & 0.6920  & 0.6724 & 0.6716  & 0.7124 & 0.5872          & 0.7684             & \textbf{0.8118} \\
			& Precision & 0.8677  & 0.8618  & 0.8611 & 0.8649  & 0.8754 & 0.8394          & 0.8908             & \textbf{0.9086} \\
			& F1-score  & 0.8524  & 0.8538  & 0.8484 & 0.8501  & 0.8657 & 0.8158          & 0.8875             & \textbf{0.9072} \\
			& Recall    & 0.8376  & 0.8460  & 0.8362 & 0.8358  & 0.8562 & 0.7936          & 0.8842             & \textbf{0.9059} \\ \midrule
			\multirow{4}{*}{WN18RR}    & Accuracy  & 0.6734  & 0.6474  & 0.6648 & 0.6453  & 0.5988 & 0.7161          & 0.6874             & \textbf{0.7315} \\
			& Precision & 0.8679  & 0.8582  & 0.8674 & 0.8605  & 0.8049 & \textbf{0.8764} & 0.8505             & 0.8694          \\
			& F1-score  & 0.8520  & 0.8406  & 0.8474 & 0.8412  & 0.8022 & 0.8672          & 0.8471             & \textbf{0.8676} \\
			& Recall    & 0.8367  & 0.8237  & 0.8282 & 0.8227  & 0.7994 & 0.8581          & 0.8437             & \textbf{0.8658} \\ \midrule
			\multirow{4}{*}{FB15K-237} & Accuracy  & 0.5541  & 0.6594  & 0.6990 & 0.7312  & 0.7635 & 0.6443          & \multirow{4}{*}{—} & \textbf{0.9593} \\
			& Precision & 0.8435  & 0.8439  & 0.8721 & 0.8829  & 0.8954 & 0.8501          &                    & \textbf{0.9801} \\
			& F1-score  & 0.8089  & 0.8367  & 0.8606 & 0.8742  & 0.8885 & 0.8307          &                    & \textbf{0.9798} \\
			& Recall    & 0.7771  & 0.8297  & 0.8495 & 0.8656  & 0.8818 & 0.8122          &                    & \textbf{0.9797} \\ \midrule
			\multirow{4}{*}{Kinship}   & Accuracy  & 0.5857  & 0.5912  & 0.6387 & 0.6508  & 0.6294 & 0.6797          & 0.7011             & \textbf{0.8030} \\
			& Precision & 0.8370  & 0.8452  & 0.8518 & 0.8587  & 0.8566 & 0.8686          & 0.8792             & \textbf{0.9341} \\
			& F1-score  & 0.8143  & 0.8197  & 0.8353 & 0.8417  & 0.8351 & 0.8540          & 0.8646             & \textbf{0.9309} \\
			& Recall    & 0.7928  & 0.7956  & 0.8194 & 0.8254  & 0.8147 & 0.8399          & 0.8506             & \textbf{0.9278} \\ \midrule
			\multirow{4}{*}{UMLS}      & Accuracy  & 0.6399  & 0.7141  & 0.7640 & 0.7973  & 0.8033 & 0.8185          & 0.8201             & \textbf{0.9228} \\
			& Precision & 0.8649  & 0.8888  & 0.9022 & 0.9132  & 0.9154 & 0.9209          & 0.9228             & \textbf{0.9640} \\
			& F1-score  & 0.8419  & 0.8726  & 0.8919 & 0.9058  & 0.9085 & 0.9150          & 0.9163             & \textbf{0.9626} \\
			& Recall    & 0.8200  & 0.8570  & 0.8820 & 0.8986  & 0.9017 & 0.9092          & 0.9100             & \textbf{0.9614} \\ \midrule
			\multirow{4}{*}{Nations}   & Accuracy  & 0.6219  & 0.5771  & 0.5323 & 0.6020  & 0.5871 & 0.7214          & 0.6816             & \textbf{0.7861} \\
			& Precision & 0.8485  & 0.8200  & 0.8329 & 0.8512  & 0.8410 & 0.8806          & 0.8614             & \textbf{0.9026} \\
			& F1-score  & 0.8293  & 0.8039  & 0.7982 & 0.8253  & 0.8165 & 0.8705          & 0.8509             & \textbf{0.8978} \\
			& Recall    & 0.8109  & 0.7886  & 0.7662 & 0.8010  & 0.7935 & 0.8607          & 0.8408             & \textbf{0.8930} \\ \bottomrule
		\end{tabular}%
	}
	\label{tab3}
\end{table*}
\subsection{Baselines}
Owing to the lack of existing baseline methods for the outdated fact detection task, we compared the following  KGE methods with our algorithm because the input of the detection module depends on the features generated by knowledge graph representation learning. 
\begin{itemize}
	\item \emph{DisMult}~\cite{DBLP:journals/corr/YangYHGD14a}: DisMult learns the embedding of entities and relations by employing a simplified bilinear scoring function. Furthermore, it can extract logical rules through the learned relation representation and capture the relational semantics.
	\item \emph{ComplEx}~\cite{DBLP:conf/icml/TrouillonWRGB16}: ComplEx firstly introduces complex vector space to knowledge graph embedding. Inspired by DisMult~\cite{DBLP:journals/corr/YangYHGD14a}, it can capture various binary relations, including both asymmetric and symmetric relations. 
	\item \emph{RotatE}~\cite{c4}: RotatE obtains the low dimensional feature representations by modeling three relation patterns as rotations in complex embedding space. Furthermore, it develops a novel self-adversarial negative sampling technique for a more effective training procedure.  
	\item \emph{pRotatE}: pRotatE is a variant of RotatE. It imposes a restriction on the modulus of the entity representations and only retains phase information compared with RotatE. 
	\item \emph{PairRE}~\cite{DBLP:conf/acl/ChaoHWC20}: PairRE is a distance-based knowledge graph embedding approach, which is simultaneously capable of dealing with complex relations and tackling three relation patterns.
	\item \emph{CompGCN}~\cite{c3}: CompGCN is a  generalized architecture utilizing the graph convolution networks technique. It can alleviate the issue of over-parameterization. 
	\item \emph{KBGAT}~\cite{DBLP:conf/acl/NathaniCSK19}: KBGAT is a novel attention-based method for preserving local neighbors’ hidden information for each fact in the knowledge graph. It can also aggregate information from the multi-hop neighborhood.
\end{itemize}           
Note that the above KGE methods cannot be directly applied to address our outdated fact problem. Hence, after the above techniques are trained to obtain valid entity and relationship features, we connect each baseline method with our framework's classification model composed of fully connected layers to perform the outdated fact detection task.

\subsection{Experimental Settings}
\textbf{Evaluation Metrics.} In this work, the outdated fact detection task is defined as a binary classification issue. In other words, for each triple $\langle e_{h}, r, e_{t}\rangle$ is either outdated or not outdated. Consequently, we employ four common evaluation metrics for the detection tasks, which encompass Accuracy, Precision, F1-score, and  Recall. Four classification metrics are calculated as follows:
\begin{itemize}
	\item$Accuracy=\frac{TP+TN}{TP+FP+FN+TN}$
	\item $Precision =\frac{TP}{TP+FP}$
	\item $F1-score=2 * \frac{ Precision*Recall }{ Precision +Recall }$
	\item $Recall =\frac{TP}{TP+FN}$
\end{itemize}
Specifically, $TN$ covers all the triples correctly classified as outdated facts, and $TP$ denotes the set of correctly classified as non-outdated facts. $FN$ and $FP$ represent sets that are incorrectly classified as outdated and non-outdated facts, respectively.

\textbf{Parameter settings.} For the consideration of the performance and efficiency of our outdated fact detection framework DEAN, we utilize a two-layers graph attention network and one contrastive R2N module for feature learning in the experiment. For attention layers, the head number is set to 2. For the contrastive R2N module, the number of GCN layers is set to 2 since using deeper GCN ~\cite{DBLP:conf/kdd/LiuGJ20} does not work well. The embedding dimension of entities and relationships is set to 200. In the training phase of this part, the batch size is set to 128 for each dataset. The Adam optimization is applied for training. We train the feature learning model for each dataset with 100 epochs. The learning rate is set to 1e-3. A hyperparameter $\lambda$ here is set to 1.0. 

\subsection{Outdated Fact Detection Results}
In this subsection, we evaluate the outdated fact detection efficiency and performance of DEAN compared with the baseline algorithms. The results of the four metrics on the six datasets are presented in \tablename{~\ref{tab3}}, where the bold values indicate the optimal performance. According to the experimental results, we observe that our framework outperforms all contrasting methods on five datasets across four metrics. In WN18RR, our method result is lower than the baselines only in terms of Precision. The experimental results demonstrate the effectiveness of our framework. Compared to the knowledge graph embedding methods, the distance-based and semantic-based, we  observe from  \tablename{~\ref{tab3}} that DEAN achieves better performance on the outdated fact detection task. This observation demonstrates that both the graph structure characteristics and attributes comprise worthwhile information for outdated fact detection and illustrates the importance of capturing the implicit structural information. Compared with GNN-based KGE methods, KBGAT, etc., we also observe that DEAN improves effectiveness on our proposed task due to the designed contrastive R2N module to enhance the learning of relational features.

In addition, from \tablename{~\ref{tab3}}, we observe that DEAN is more effective on the FB15K-237, Kinship, and UMLS datasets than on the WN18, WN18RR, and Nations. Compared to all baselines, DEAN achieves an improvement of approximately $10\%$ in terms of accuracy on the FB15K-237, Kinship, and UMLS datasets. In addition, we found that the number of relationship types in these three datasets is significantly greater than those in the remaining three. It can be seen that DEAN performs better with an increased number of relationships. The explanation for this is that a contrastive R2N module is designed in our framework to strengthen relational learning features, which illustrates more expressive relationships making our model more effective on the outdated fact detection task. Note that the detection effectiveness for the Nations dataset is not very obvious. The performance is limited by the small number of entity types and facts. 
\subsection{Parameter Sensitivity Analysis}
In this subsection, we conduct experiments to investigate the importance of three crucial hyperparameters on the performance of DEAN, including the number of heads, the coefficient of the loss function, and the embedding dimension.
\begin{figure}[t]
	\centering
	\subfigure[WN18]{
		\begin{minipage}[t]{0.5\linewidth}
			\centering
			\label{fig4a}
			\includegraphics[width=1.0\hsize,height=1.0\hsize]{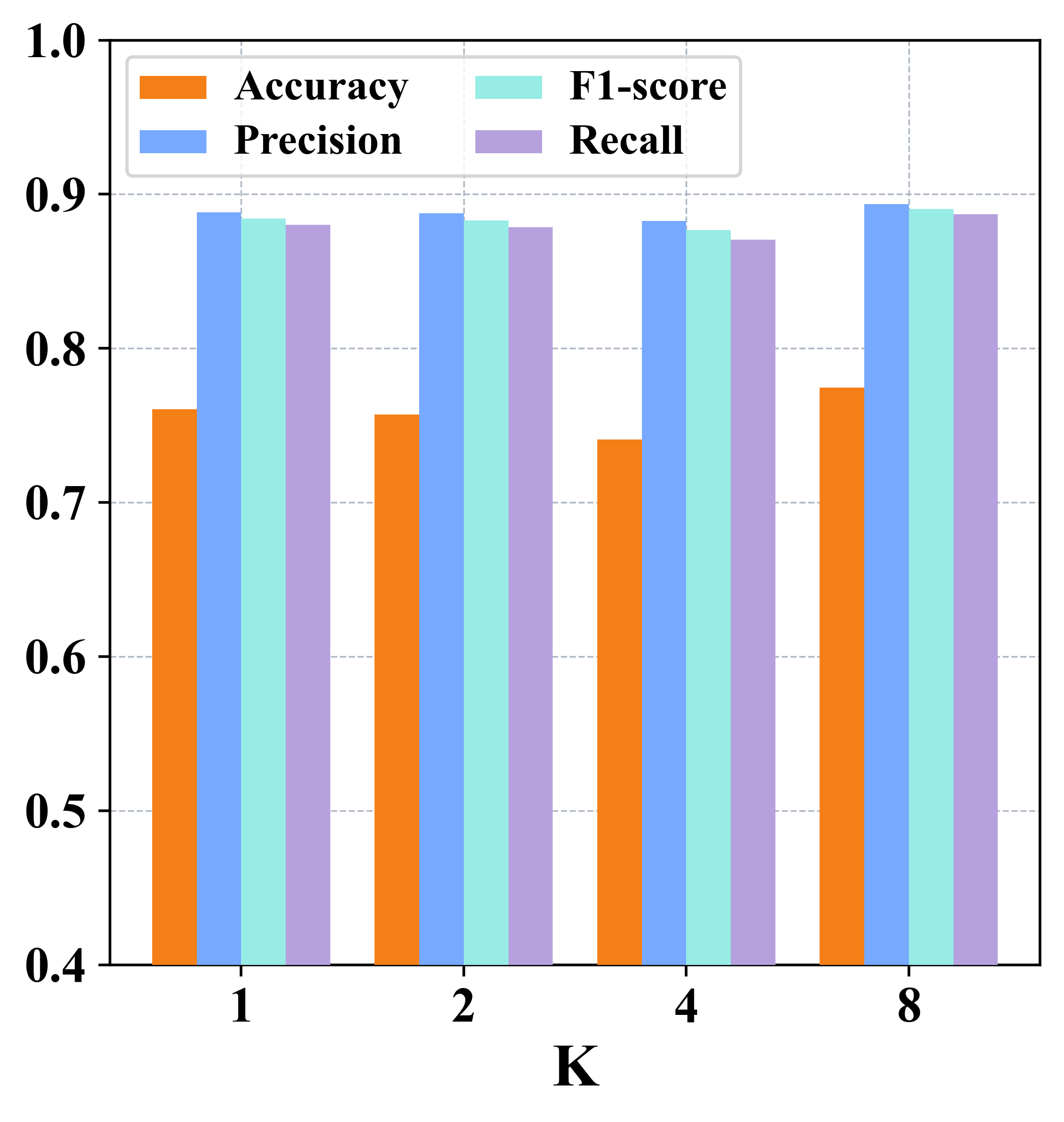}
		\end{minipage}%
	}%
	\subfigure[WN18RR]{
		\begin{minipage}[t]{0.5\linewidth}
			\centering
			\label{fig4b}
			\includegraphics[width=1.0\hsize,height=1.0\hsize]{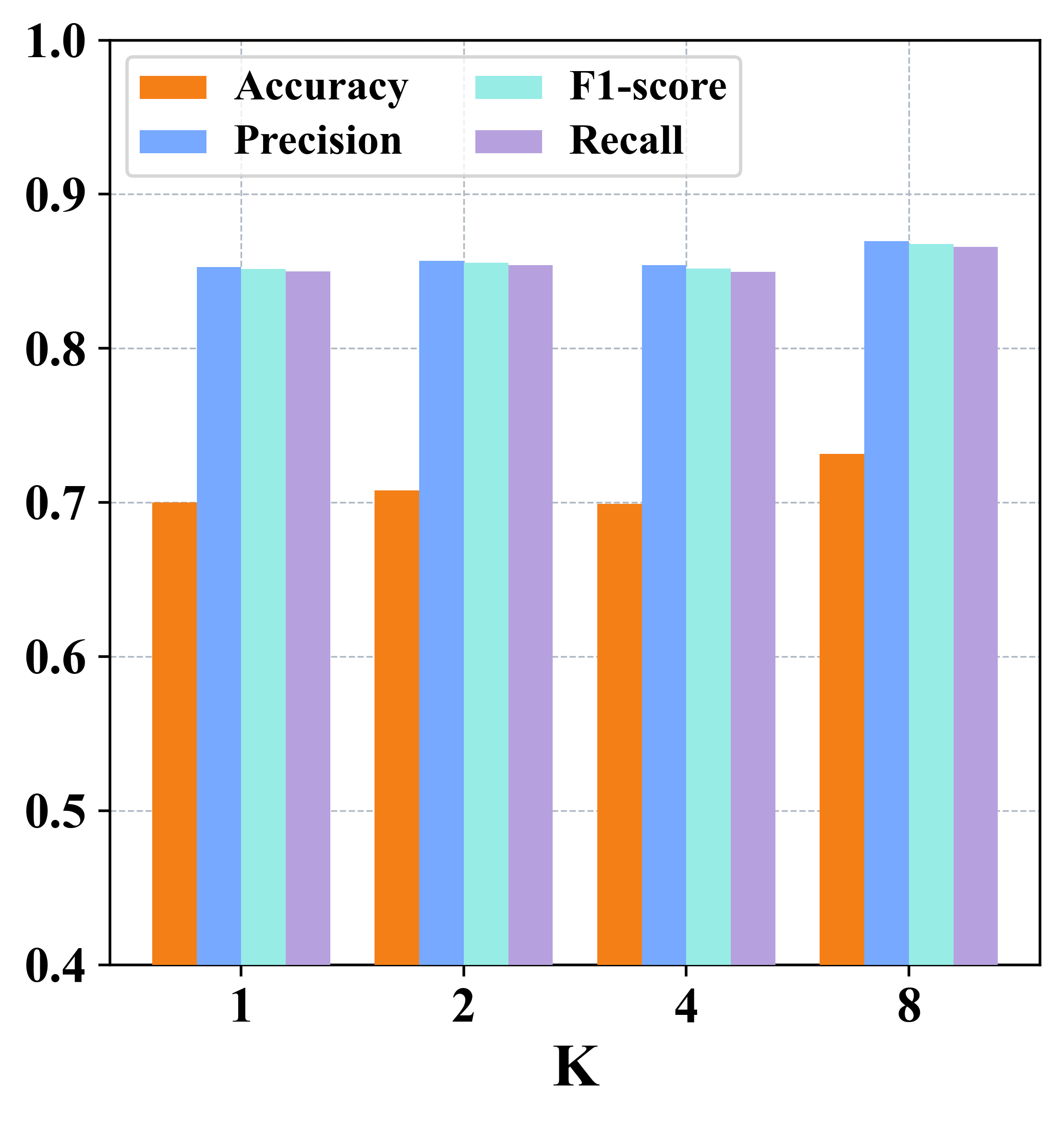}
		\end{minipage}%
	}%
	
	\subfigure[FB15K-237]{
		\begin{minipage}[t]{0.5\linewidth}
			\centering
			\label{fig4c}
			\includegraphics[width=1.0\hsize,height=1.0\hsize]{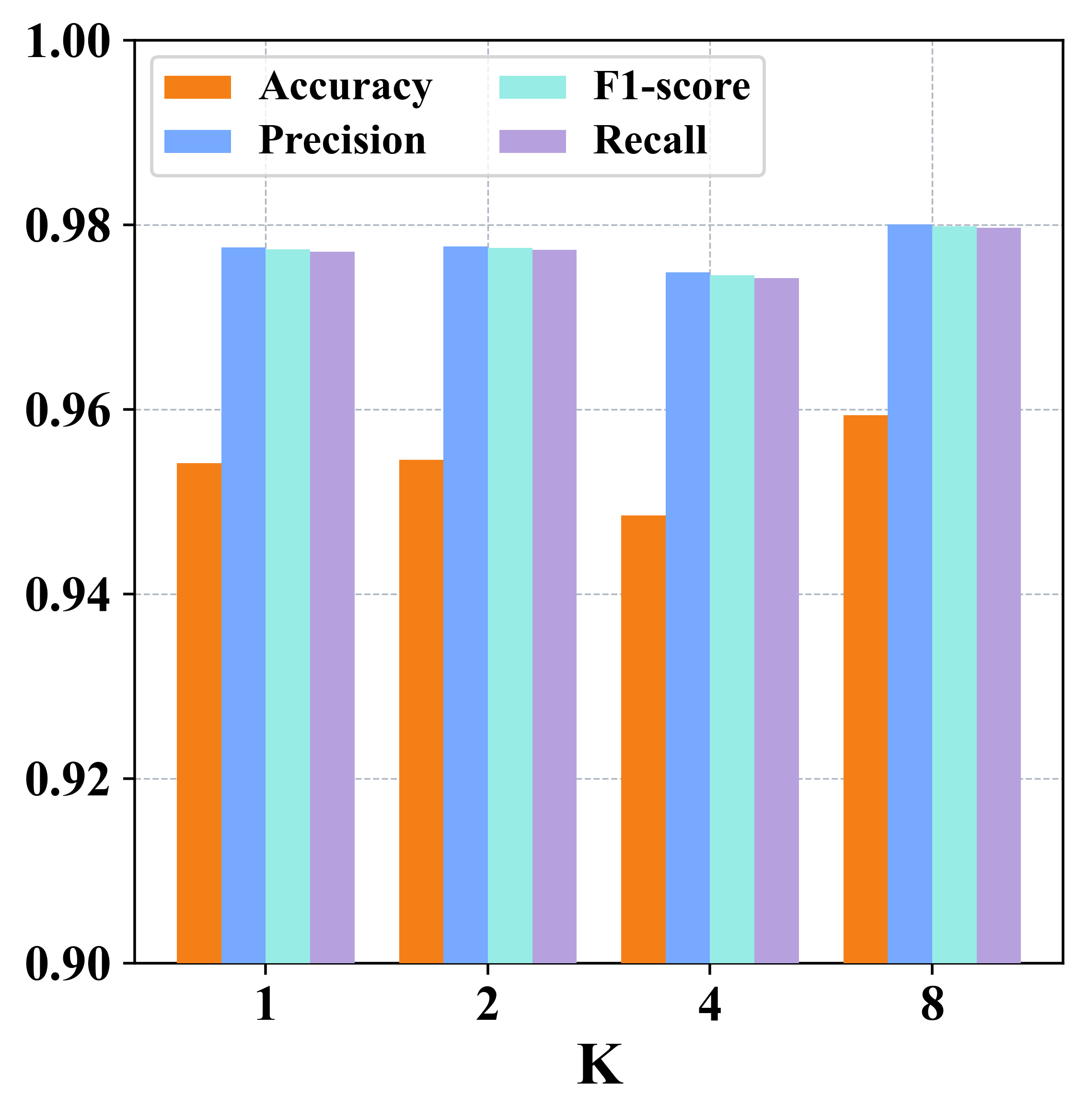}
		\end{minipage}%
	}%
	\subfigure[Kinship]{
		\begin{minipage}[t]{0.5\linewidth}
			\centering
			\label{fig4d}
			\includegraphics[width=1.0\hsize,height=1.0\hsize]{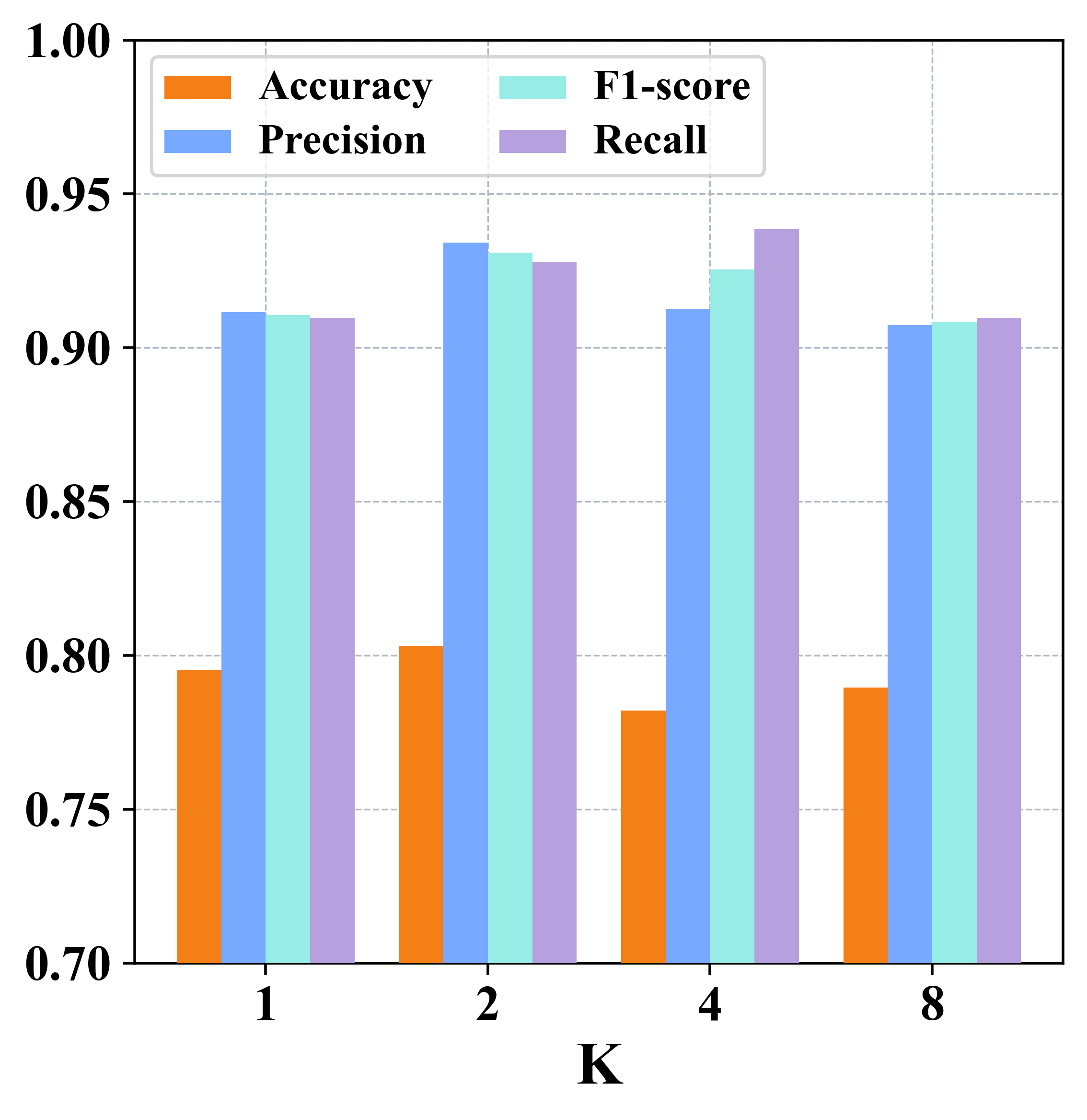}
		\end{minipage}
	}%
	
	\subfigure[UMLS]{
		\begin{minipage}[t]{0.5\linewidth}
			\centering
			\label{fig4e}
			\includegraphics[width=1.0\hsize,height=1.0\hsize]{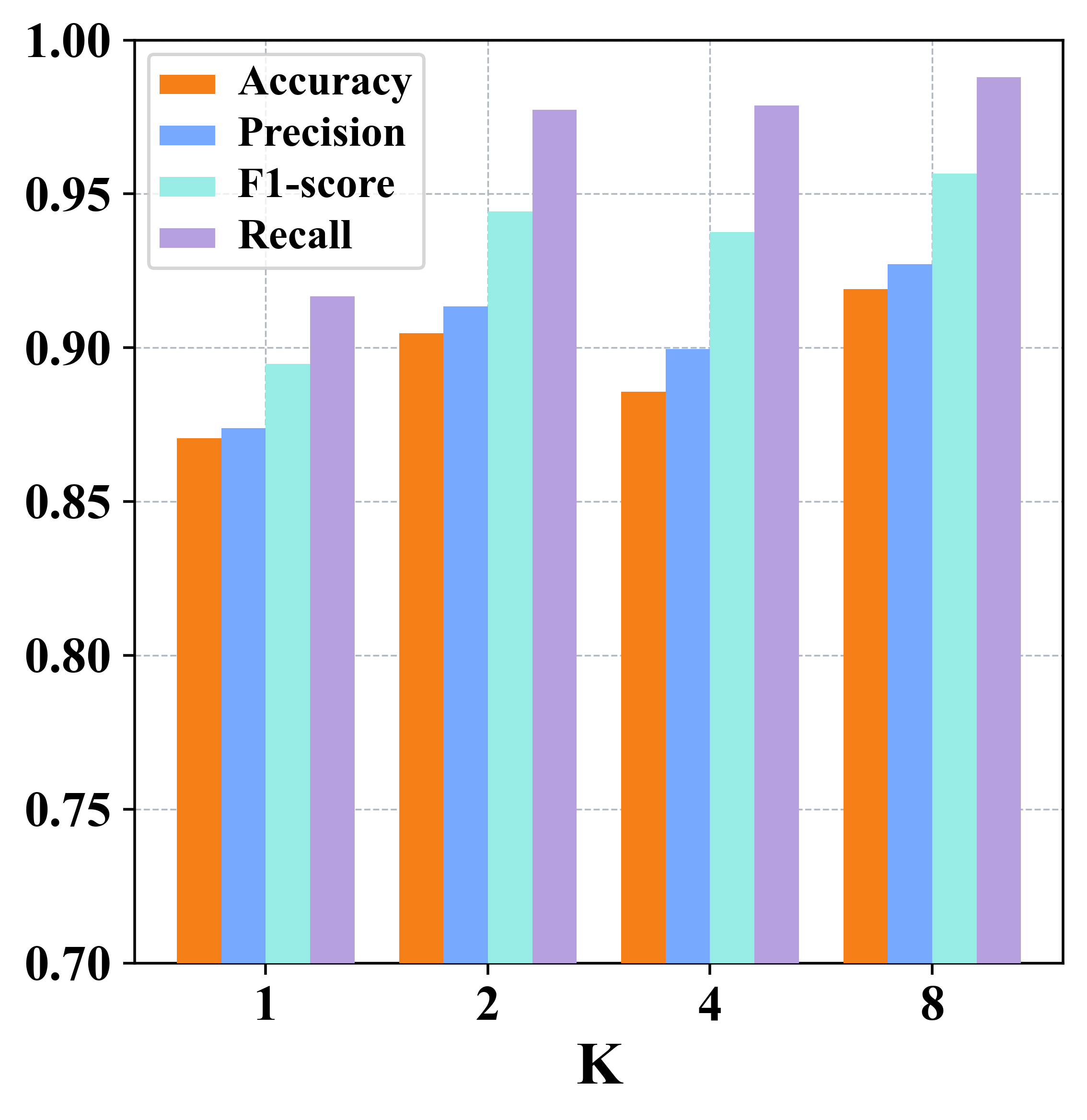}
		\end{minipage}
	}%
	\subfigure[Nations]{
		\begin{minipage}[t]{0.5\linewidth}
			\centering
			\label{fig4f}
			\includegraphics[width=1.0\hsize,height=1.0\hsize]{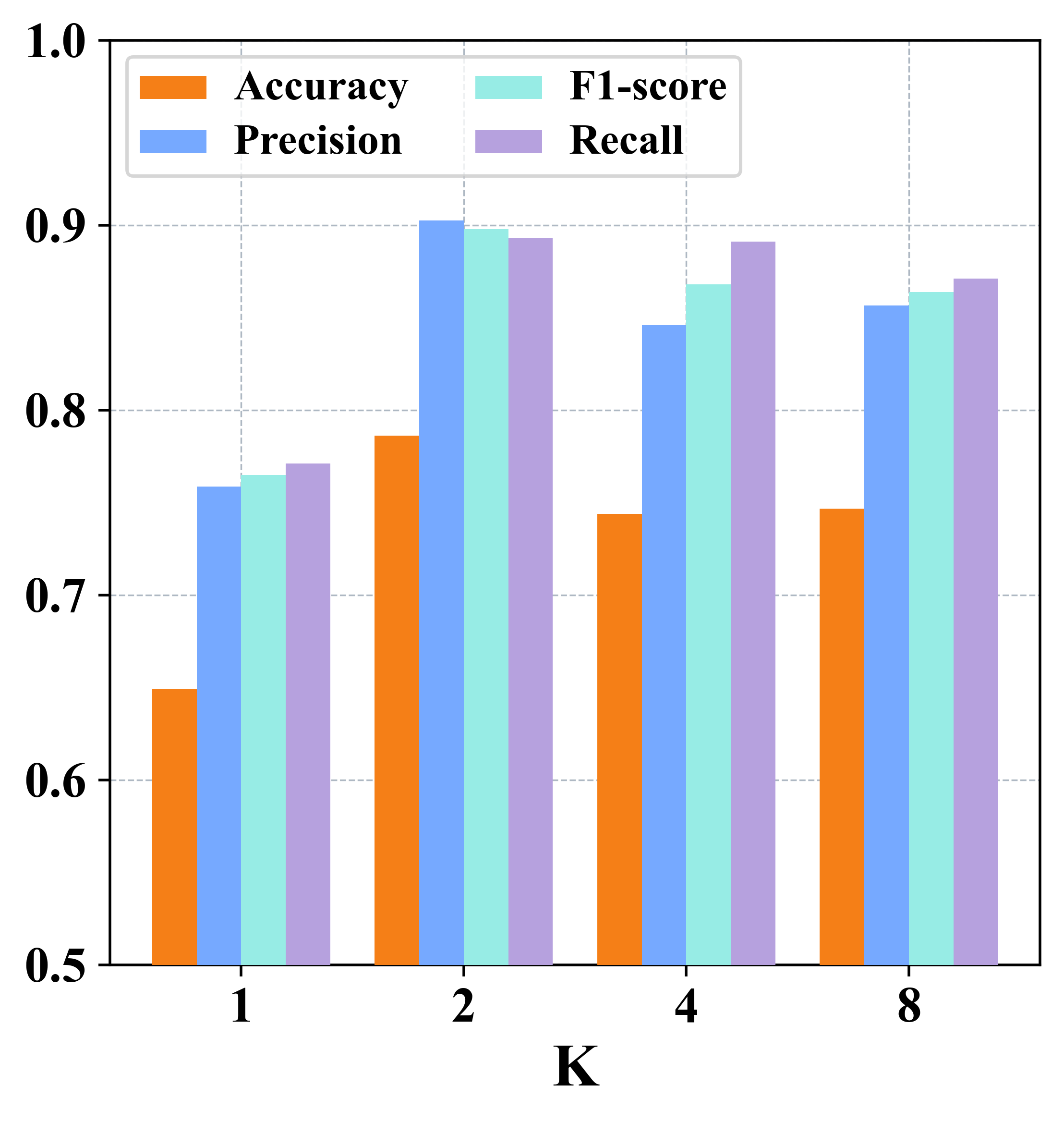}
		\end{minipage}%
	}%
	\centering
	\caption{Sensitivities of DEAN w.r.t. K in different datasets.}
	\label{fig4}
\end{figure}
\begin{figure}[t]
	\centering
	\subfigure[Accuracy]{
		\begin{minipage}[t]{0.5\linewidth}
			\centering
			\label{fig5a}
			\includegraphics[width=1.0\hsize,height=1.0\hsize]{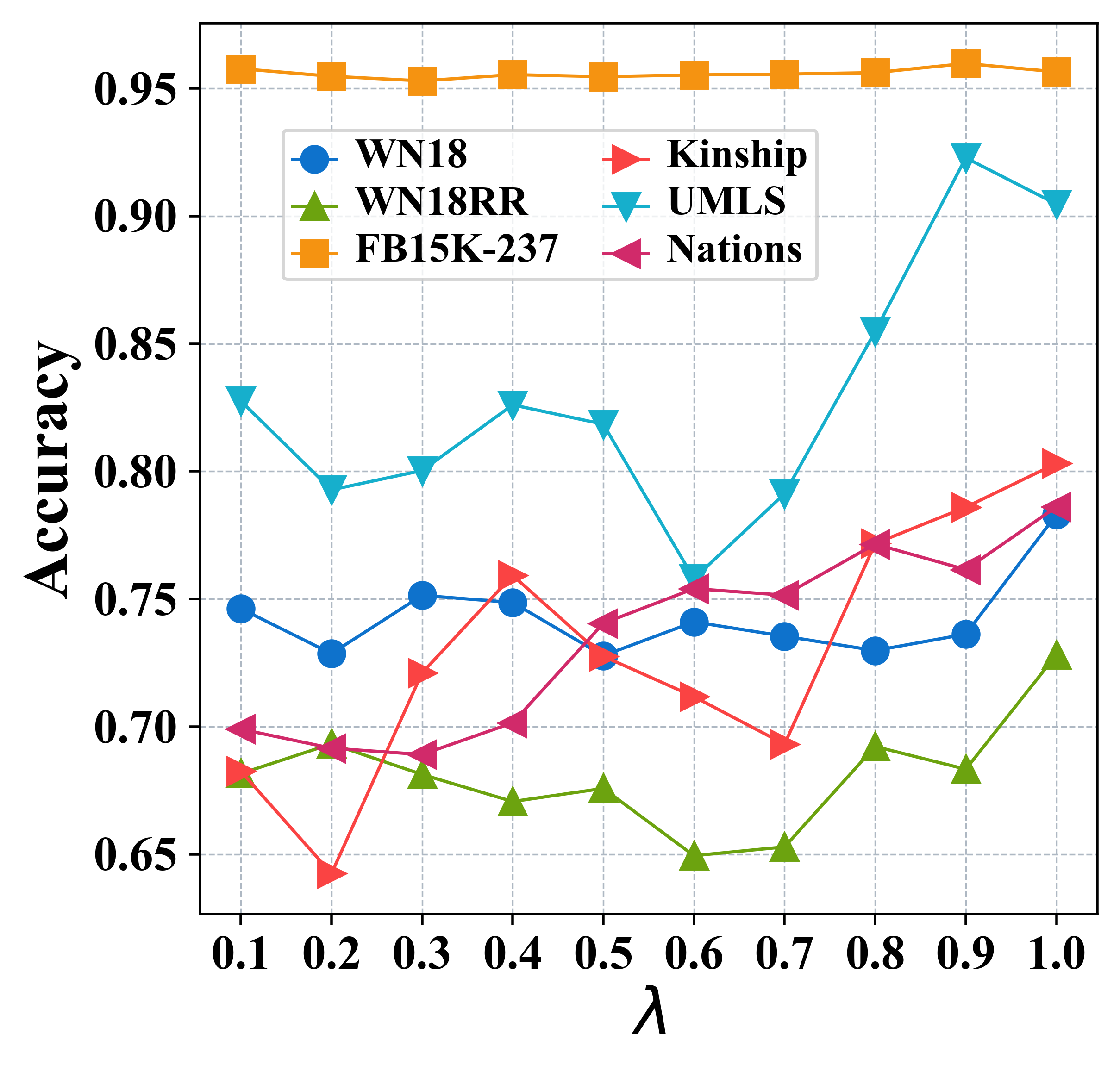}
		\end{minipage}%
	}%
	\subfigure[Precision]{
		\begin{minipage}[t]{0.5\linewidth}
			\centering
			\label{fig5b}
			\includegraphics[width=1.0\hsize,height=1.0\hsize]{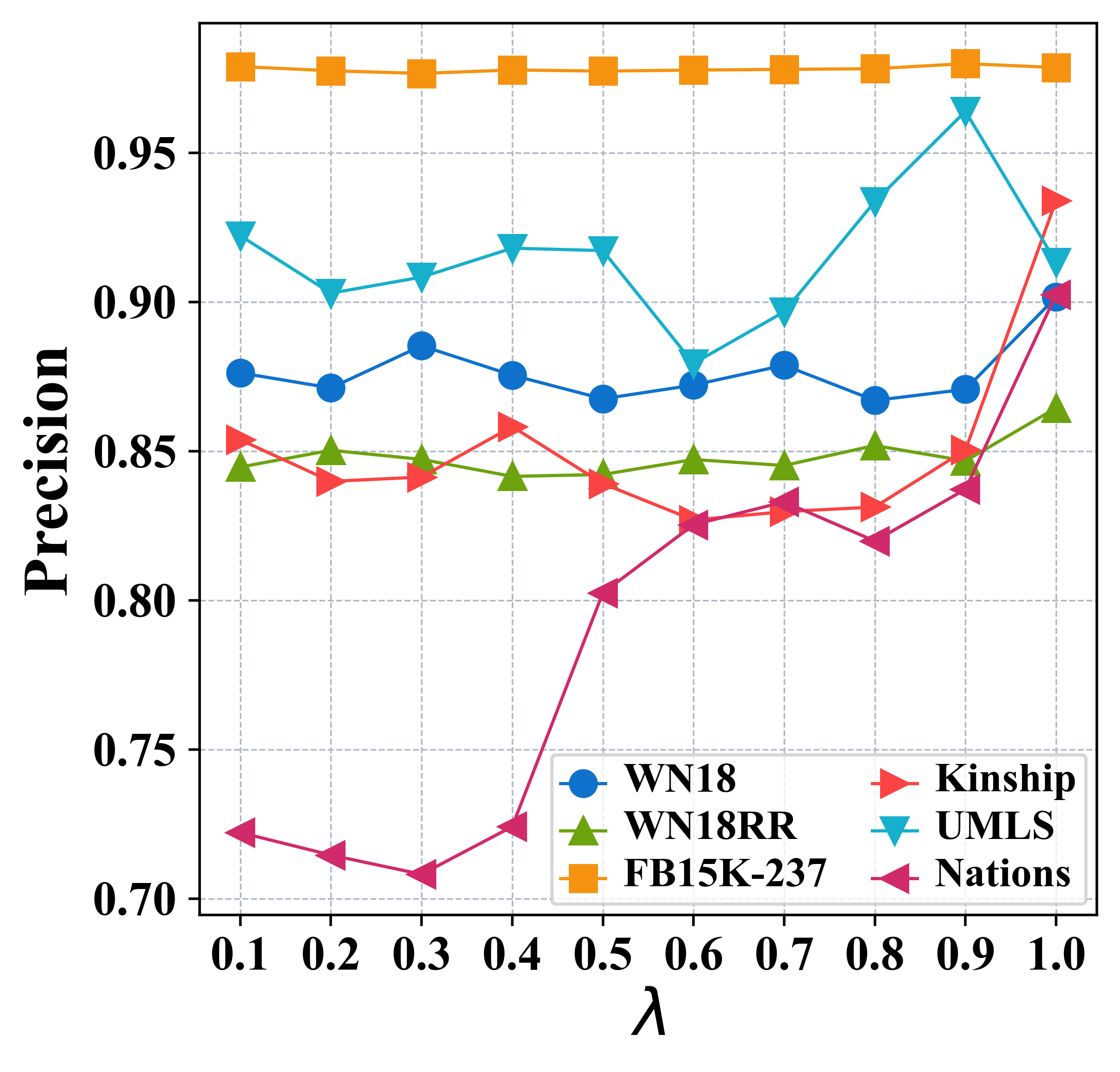}
		\end{minipage}%
	}%
	
	\subfigure[F1-score]{
		\begin{minipage}[t]{0.5\linewidth}
			\centering
			\label{fig5c}
			\includegraphics[width=1.0\hsize,height=1.0\hsize]{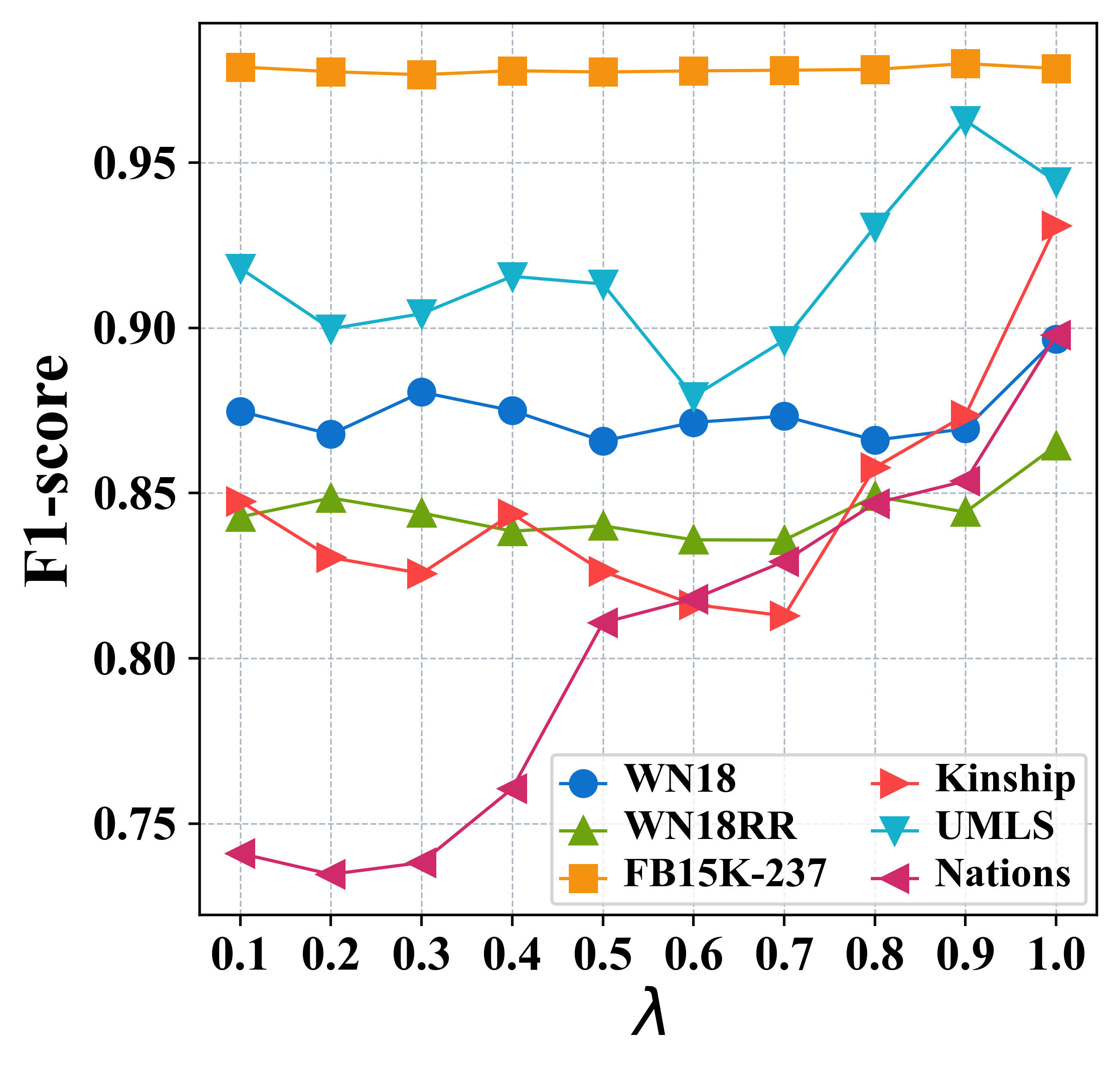}
		\end{minipage}
	}%
	\subfigure[Recall]{
		\begin{minipage}[t]{0.5\linewidth}
			\centering
			\label{fig5d}
			\includegraphics[width=1.0\hsize,height=1.0\hsize]{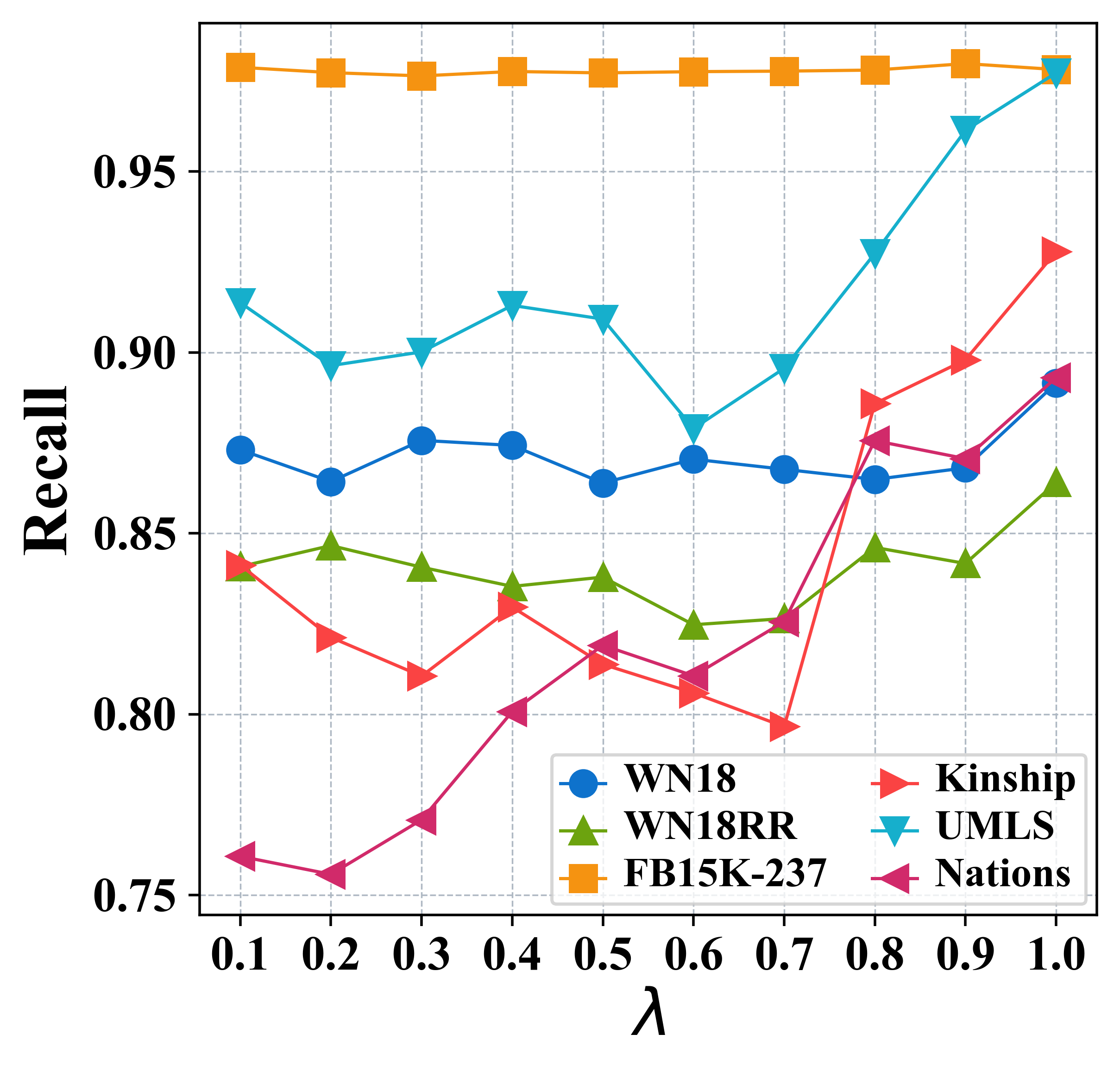}
		\end{minipage}
	}%
	\centering
	\caption{ Sensitivities of DEAN w.r.t. $\lambda$}
	\label{fig5}
\end{figure}

\textbf{Number of heads.} In this experiment, we alter the head values $K$ to investigate its influence on four different evaluation metrics. Figure~\ref{fig4} illustrates the performance of DEAN when the number of head $K$ is set for the exponential growth from 1 to 8. As we can see from Figure~\ref{fig4a} to Figure~\ref{fig4c}, DEAN changes relatively stably on all metrics as the value of $K$ is set differently. However, from Figure~\ref{fig4d} to Figure~\ref{fig4f}, DEAN achieves better performance when multi-head attention is applied. This observation illustrates that the extension of the multi-head attention mechanism under such a task situation needs to be adjusted according to the characteristics of different datasets.

\textbf{Coefficient of loss function.} We explore the effect of coefficient $\lambda$ for our proposed framework DEAN. Figure~\ref{fig5} illustrates how different choices of $\lambda$ have an impact on the performance of DEAN. We select the value in the range from 0.1 to 1.0. As shown in the figure, we make the observations that DEAN achieves the optimal performance on FB15K-237 for all metrics, including Accuracy, Precision, F1-score, and Recall. In addition, for WN18, WN18RR, Kinship, and Nations, we observe that DEAN achieves better performance when $\lambda$=1.0. Due to the structural differences of entities and relationships in different datasets, our framework also achieved better performance on the remaining two datasets when the $\lambda$ were closer to 1.0. In this experiment, our designed contrastive R2N module can significantly contribute to the overall framework's performance improvement when $\lambda$ is set to 1.0. This observation demonstrates that the designed loss function $\mathcal{L}_{R2N}$ is effective.

\begin{figure}[t]
	\centering
	\subfigure[Accuracy]{
		\begin{minipage}[t]{0.5\linewidth}
			\centering
			\label{fig6a}
			\includegraphics[width=1.0\hsize,height=1.0\hsize]{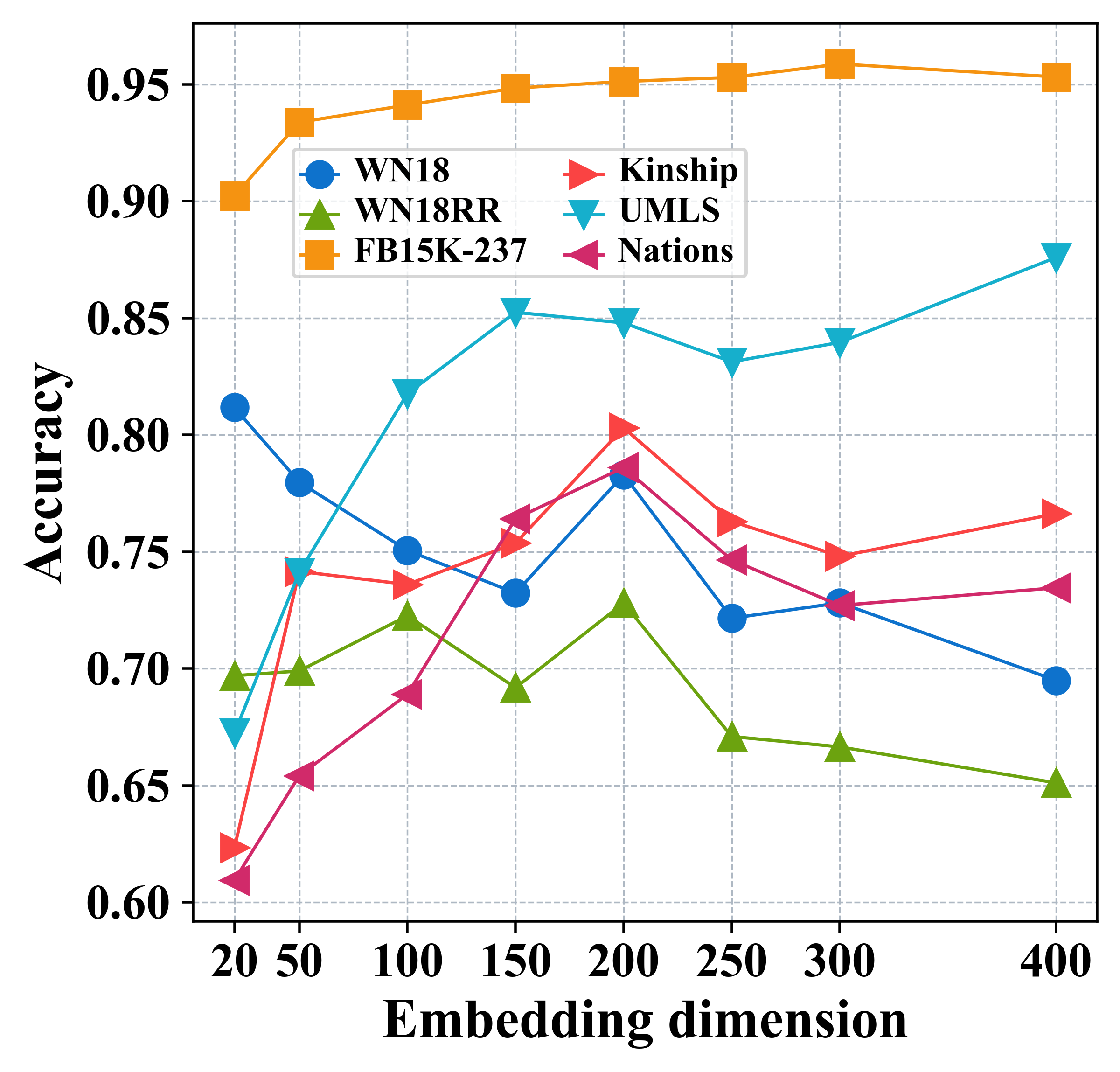}
		\end{minipage}%
	}%
	\subfigure[Precision]{
		\begin{minipage}[t]{0.5\linewidth}
			\centering
			\label{fig6b}
			\includegraphics[width=1.0\hsize,height=1.0\hsize]{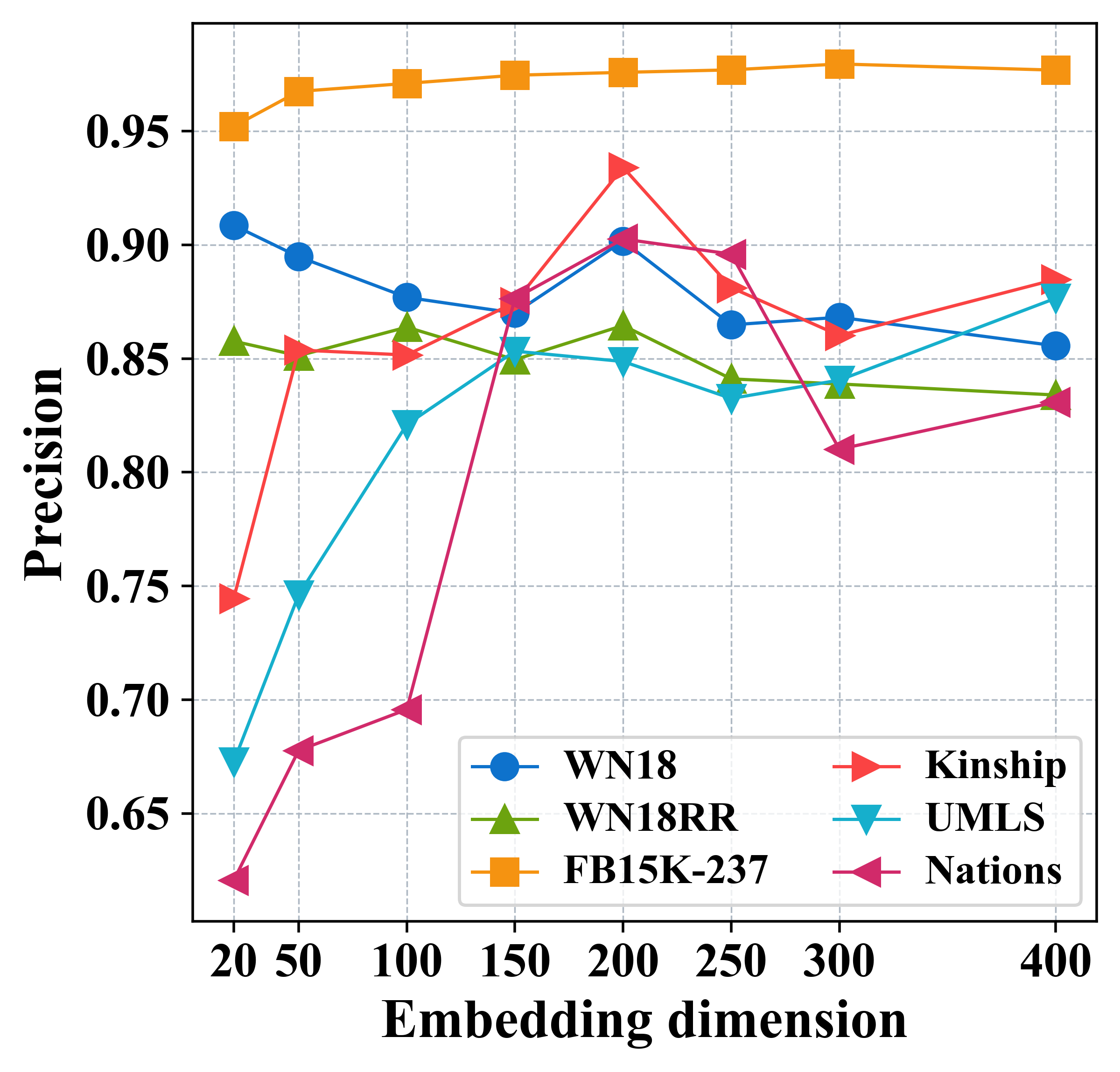}
		\end{minipage}%
	}%
	
	\subfigure[F1-score]{
		\begin{minipage}[t]{0.5\linewidth}
			\centering
			\label{fig6c}
			\includegraphics[width=1.0\hsize,height=1.0\hsize]{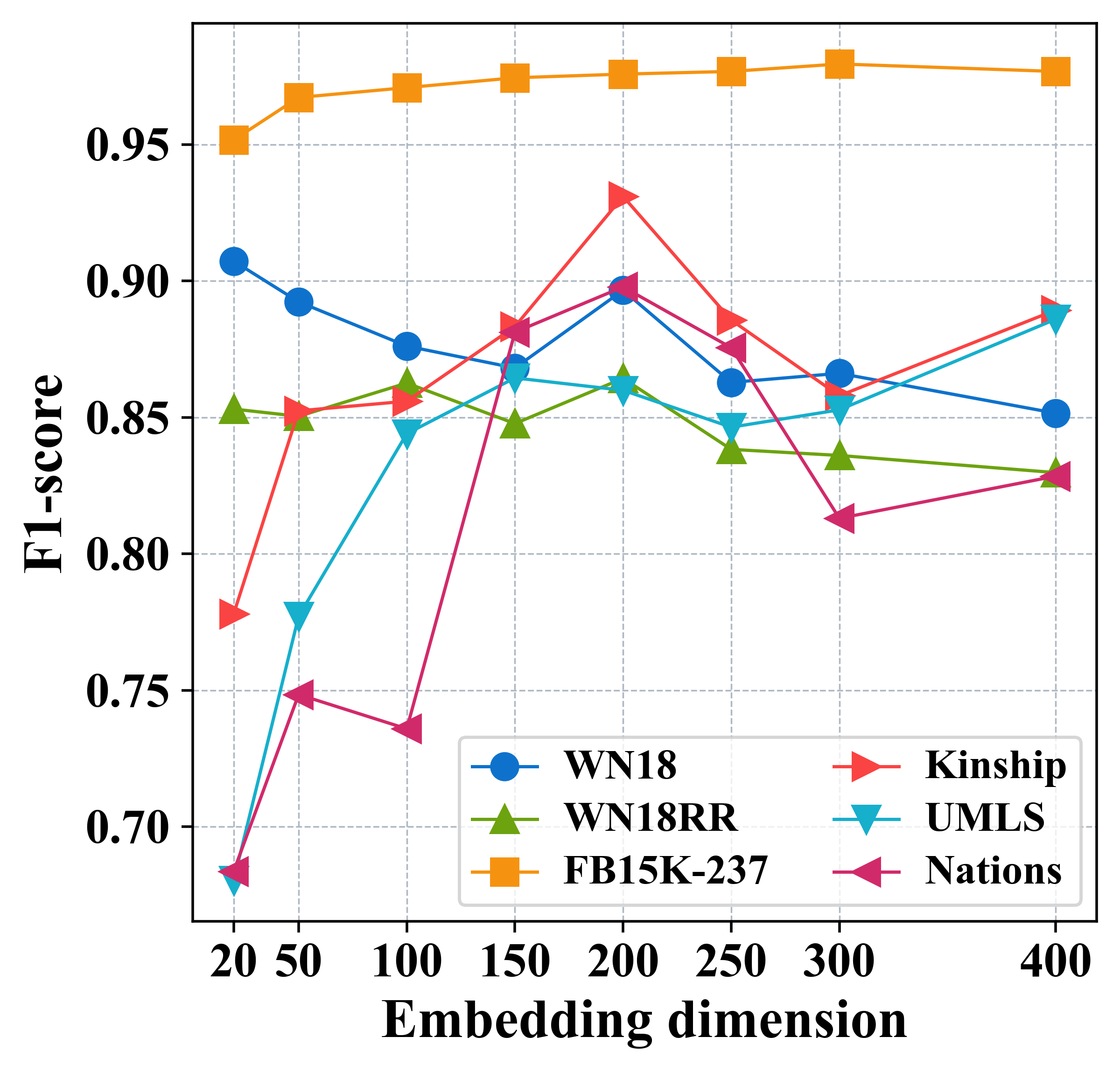}
		\end{minipage}
	}%
	\subfigure[Recall]{
		\begin{minipage}[t]{0.5\linewidth}
			\centering
			\label{fig6d}
			\includegraphics[width=1.0\hsize,height=1.0\hsize]{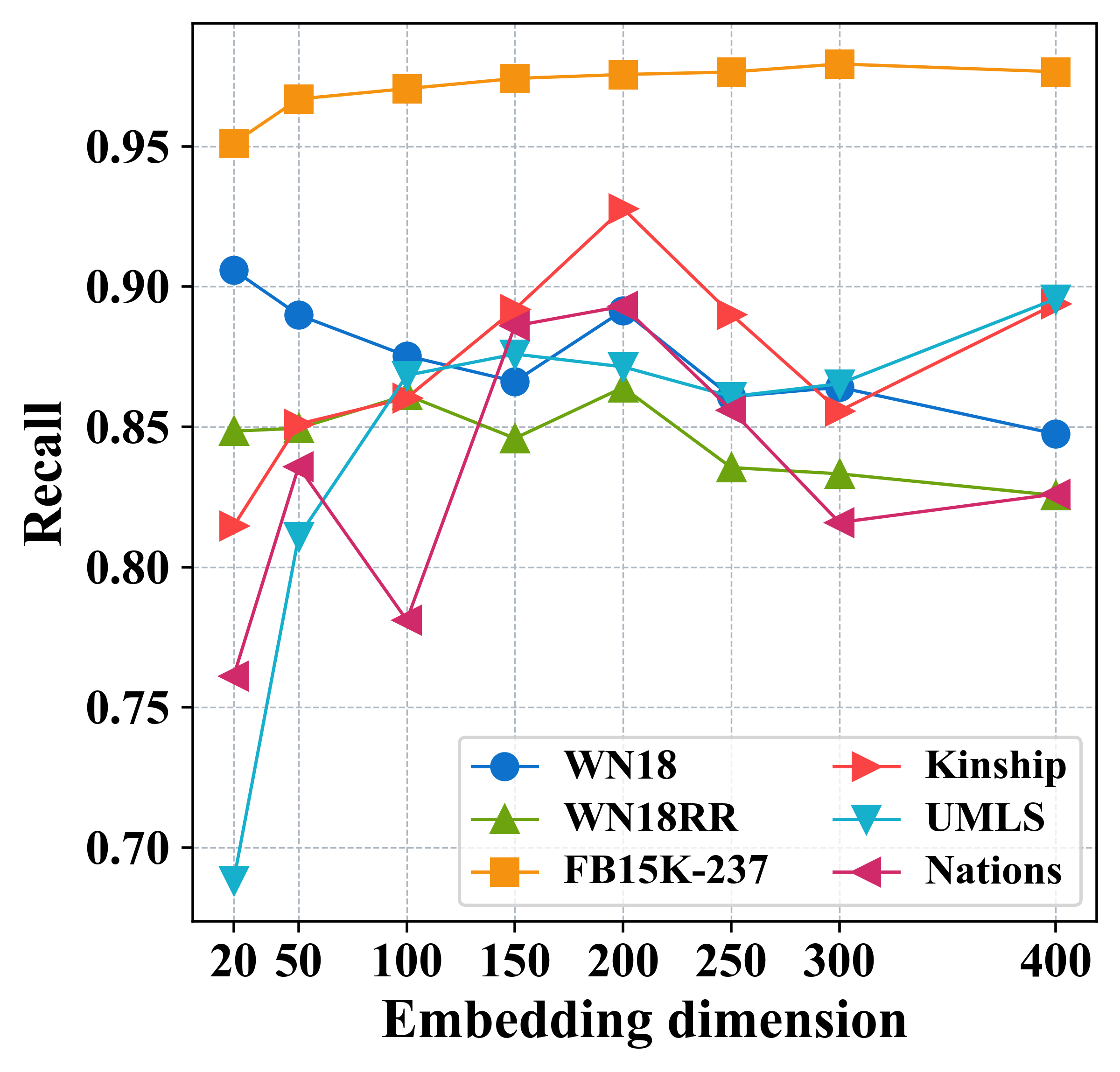}
		\end{minipage}
	}%
	\centering
	\caption{ Sensitivities of DEAN w.r.t. $d$}
	\label{fig6}
\end{figure}

\textbf{Embedding dimension.} We further study the significance of the embedding dimension \emph{d}. The performance change of DEAN is demonstrated in Figure~\ref{fig6}. We modify the values of the embedding dimension from 20 to 400. For the FB15K-237 dataset, the performance of DEAN on all metrics increases slowly with a larger embedding dimension. Moreover, as we can see in Figure~\ref{fig6}, for the WN18RR and WN18 datasets, the performance of outdated fact detection framework decreased slightly with the change of embedding dimension. For the remaining three datasets, the performance of DEAN  fluctuates significantly but the best results are achieved when the dimension is set to 200. We can make the observation that there is plenty of helpful information for the outdated fact detection task when the embedding dimension is set to 200.

\section{Conclusion}
\label{conclusion}
In this work, we have proposed DEAN, a novel deep learning based framework tailored for the identification of outdated facts within KGs. DEAN adeptly captures implicit structural information by modeling both entities and relations. To enhance the discrimination between outdated and non-outdated facts, we innovatively designed a contrastive learning method based on the pre-defined weighted R2N graph. DEAN demonstrates a unique capability to capture relation features, providing a more expressive representation in KGs.

Our extensive experiments affirm the robust performance and superiority of DEAN. Notably, it excels in scenarios where there is a diverse range of relation types. However, it's important to acknowledge that DEAN may not be optimal for large datasets with few relation types. A promising avenue for future research involves developing a more expressive and generalized solution tailored for datasets with few(er) relationships.

Furthermore, our work opens up intriguing possibilities for extending outdated fact detection beyond changes in relations to encompass changes in entities. Addressing this dimension could significantly enhance the comprehensiveness of our approach, contributing to the evolving landscape of KG maintenance and quality assurance. 

\bibliographystyle{IEEEtran}
\bibliography{IEEEabrv,DEAN.bib}

\begin{thebibliography}{10}
\providecommand{\url}[1]{#1}
\csname url@samestyle\endcsname
\providecommand{\newblock}{\relax}
\providecommand{\bibinfo}[2]{#2}
\providecommand{\BIBentrySTDinterwordspacing}{\spaceskip=0pt\relax}
\providecommand{\BIBentryALTinterwordstretchfactor}{4}
\providecommand{\BIBentryALTinterwordspacing}{\spaceskip=\fontdimen2\font plus
\BIBentryALTinterwordstretchfactor\fontdimen3\font minus
  \fontdimen4\font\relax}
\providecommand{\BIBforeignlanguage}[2]{{%
\expandafter\ifx\csname l@#1\endcsname\relax
\typeout{** WARNING: IEEEtran.bst: No hyphenation pattern has been}%
\typeout{** loaded for the language `#1'. Using the pattern for}%
\typeout{** the default language instead.}%
\else
\language=\csname l@#1\endcsname
\fi
#2}}
\providecommand{\BIBdecl}{\relax}
\BIBdecl

\bibitem{peng2023kgsurvey}
C.~Peng, F.~Xia, M.~Naseriparsa, and F.~Osborne, ``Knowledge graphs:
  Opportunities and challenges,'' \emph{Artificial Intelligence Review}, 2023.

\bibitem{ji2021survey}
S.~Ji, S.~Pan, E.~Cambria, P.~Marttinen, and S.~Y. Philip, ``A survey on
  knowledge graphs: Representation, acquisition, and applications,'' \emph{IEEE
  Transactions on Neural Networks and Learning Systems}, vol.~33, no.~2, pp.
  494--514, 2021.

\bibitem{10031054}
K.~Sun, S.~Yu, C.~Peng, X.~Li, M.~Naseriparsa, and F.~Xia, ``Abnormal
  entity-aware knowledge graph completion,'' in \emph{IEEE International
  Conference on Data Mining Workshops (ICDMW)}, 2022, pp. 891--900.

\bibitem{tran2013automatic}
T.~Tran and T.~H. Cao, ``Automatic detection of outdated information in
  wikipedia infoboxes.'' \emph{Res. Comput. Sci.}, vol.~70, pp. 211--222, 2013.

\bibitem{hao2020outdated}
S.~Hao, C.~Chai, G.~Li, N.~Tang, N.~Wang, and X.~Yu, ``Outdated fact detection
  in knowledge bases,'' in \emph{IEEE 36th International Conference on Data
  Engineering (ICDE)}, 2020, pp. 1890--1893.

\bibitem{liu2022mirror}
J.~Liu, F.~Xia, J.~Ren, B.~Xu, G.~Pang, and L.~Chi, ``Mirror: Mining implicit
  relationships via structure-enhanced graph convolutional networks,''
  \emph{ACM Transactions on Knowledge Discovery from Data (TKDD)}, vol.~17,
  no.~4, 2023.

\bibitem{PU2023104464}
Y.~Pu, D.~Beck, and K.~Verspoor, ``Graph embedding-based link prediction for
  literature-based discovery in alzheimer’s disease,'' \emph{Journal of
  Biomedical Informatics}, vol. 145, p. 104464, 2023.

\bibitem{nguyen2022node}
D.~Q. Nguyen, V.~Tong, D.~Phung, and D.~Q. Nguyen, ``Node co-occurrence based
  graph neural networks for knowledge graph link prediction,'' in
  \emph{Proceedings of the ACM International Conference on Web Search and Data
  Mining}, 2022, pp. 1589--1592.

\bibitem{liu2022selfkg}
X.~Liu, H.~Hong, X.~Wang, Z.~Chen, E.~Kharlamov, Y.~Dong, and J.~Tang,
  ``Selfkg: Self-supervised entity alignment in knowledge graphs,'' in
  \emph{Proceedings of the ACM Web Conference 2022}, 2022, pp. 860--870.

\bibitem{DBLP:conf/iclr/VelickovicCCRLB18}
P.~Velickovic, G.~Cucurull, A.~Casanova, A.~Romero, P.~Li{\`{o}}, and
  Y.~Bengio, ``Graph attention networks,'' in \emph{Proceedings of the
  International Conference on Learning Representations (ICLR)}, 2018.

\bibitem{6719874}
T.~Tran and T.~H. Cao, ``A hybrid method for detecting outdated information in
  wikipedia infoboxes,'' in \emph{The 2013 RIVF International Conference on
  Computing Communication Technologies - Research, Innovation, and Vision for
  Future (RIVF)}, 2013, pp. 97--102.

\bibitem{DBLP:conf/ijcai/LiangZX17}
J.~Liang, S.~Zhang, and Y.~Xiao, ``How to keep a knowledge base synchronized
  with its encyclopedia source,'' in \emph{Proceedings of the International
  Joint Conference on Artificial Intelligence (IJCAI)}, 2017, pp. 3749--3755.

\bibitem{tang2019learning}
J.~Tang, Y.~Feng, and D.~Zhao, ``Learning to update knowledge graphs by reading
  news,'' in \emph{Proceedings of the 2019 Conference on Empirical Methods in
  Natural Language Processing and the 9th International Joint Conference on
  Natural Language Processing (EMNLP-IJCNLP)}, 2019, pp. 2632--2641.

\bibitem{DBLP:conf/kdd/HuangSDLL021}
H.~Huang, L.~Sun, B.~Du, C.~Liu, W.~Lv, and H.~Xiong, ``Representation learning
  on knowledge graphs for node importance estimation,'' in \emph{{KDD} '21: The
  27th {ACM} {SIGKDD} Conference on Knowledge Discovery and Data Mining,
  Virtual Event, Singapore, August 14-18, 2021}.\hskip 1em plus 0.5em minus
  0.4em\relax {ACM}, 2021, pp. 646--655.

\bibitem{Xia2021TAI}
F.~Xia, K.~Sun, S.~Yu, A.~Aziz, L.~Wan, S.~Pan, and H.~Liu, ``Graph learning: A
  survey,'' \emph{IEEE Transactions on Artificial Intelligence}, vol.~2, no.~2,
  pp. 109--127, 2021.

\bibitem{DBLP:conf/nips/BordesUGWY13}
A.~Bordes, N.~Usunier, A.~Garc{\'{\i}}a{-}Dur{\'{a}}n, J.~Weston, and
  O.~Yakhnenko, ``Translating embeddings for modeling multi-relational data,''
  in \emph{Advances in nformation Processing Systems (NeurIPS)}, 2013, pp.
  2787--2795.

\bibitem{wang2014knowledge}
Z.~Wang, J.~Zhang, J.~Feng, and Z.~Chen, ``Knowledge graph embedding by
  translating on hyperplanes,'' in \emph{Proceedings of the AAAI Conference on
  Artificial Intelligence}, 2014, pp. 1112--1119.

\bibitem{DBLP:conf/aaai/LinLSLZ15}
Y.~Lin, Z.~Liu, M.~Sun, Y.~Liu, and X.~Zhu, ``Learning entity and relation
  embeddings for knowledge graph completion,'' in \emph{Proceedings of the AAAI
  Conference on Artificial Intelligence}, 2015, pp. 2181--2187.

\bibitem{DBLP:conf/aaai/DettmersMS018}
T.~Dettmers, P.~Minervini, P.~Stenetorp, and S.~Riedel, ``Convolutional 2d
  knowledge graph embeddings,'' in \emph{Proceedings of the AAAI Conference on
  Artificial Intelligence}, 2018, pp. 1811--1818.

\bibitem{DBLP:conf/esws/SchlichtkrullKB18}
M.~S. Schlichtkrull, T.~N. Kipf, P.~Bloem, R.~van~den Berg, I.~Titov, and
  M.~Welling, ``Modeling relational data with graph convolutional networks,''
  in \emph{European Semantic Web Conference (ESWC)}, vol. 10843, 2018, pp.
  593--607.

\bibitem{DBLP:conf/iclr/KipfW17}
T.~N. Kipf and M.~Welling, ``Semi-supervised classification with graph
  convolutional networks,'' in \emph{Proceedings of the International
  Conference on Learning Representations (ICLR)}, 2017.

\bibitem{shang2019end}
C.~Shang, Y.~Tang, J.~Huang, J.~Bi, X.~He, and B.~Zhou, ``End-to-end
  structure-aware convolutional networks for knowledge base completion,'' in
  \emph{Proceedings of the AAAI Conference on Artificial Intelligence}, 2019,
  pp. 3060--3067.

\bibitem{DBLP:conf/acl/NathaniCSK19}
D.~Nathani, J.~Chauhan, C.~Sharma, and M.~Kaul, ``Learning attention-based
  embeddings for relation prediction in knowledge graphs,'' in
  \emph{Proceedings of the Association for Computational Linguistics (ACL)},
  2019, pp. 4710--4723.

\bibitem{yu2023spatio}
S.~Yu, F.~Xia, S.~Li, M.~Hou, and Q.~Z. Sheng, ``Spatio-temporal graph learning
  for epidemic prediction,'' \emph{ACM Transactions on Intelligent Systems and
  Technology}, vol.~14, no.~2, 2023.

\bibitem{xia2022cengcn}
F.~Xia, L.~Wang, T.~Tang, X.~Chen, X.~Kong, G.~Oatley, and I.~King, ``Cengcn:
  Centralized convolutional networks with vertex imbalance for scale-free
  graphs,'' \emph{IEEE Transactions on Knowledge and Data Engineering},
  vol.~35, no.~5, pp. 4555--4569, 2023.

\bibitem{velickovic2019deep}
P.~Velickovic, W.~Fedus, W.~L. Hamilton, P.~Li{\`o}, Y.~Bengio, and R.~D.
  Hjelm, ``Deep graph infomax.'' in \emph{Proceedings of the International
  Conference on Learning Representations (ICLR)}, 2019.

\bibitem{xie2018does}
R.~Xie, Z.~Liu, F.~Lin, and L.~Lin, ``Does william shakespeare really write
  hamlet? knowledge representation learning with confidence,'' in
  \emph{Proceedings of the AAAI Conference on Artificial Intelligence}, 2018.

\bibitem{ren2022graph}
J.~Ren, F.~Xia, I.~Lee, A.~Noori~Hoshyar, and C.~Aggarwal, ``Graph learning for
  anomaly analytics: Algorithms, applications, and challenges,'' \emph{ACM
  Transactions on Intelligent Systems and Technology}, vol.~14, no.~2, 2023.

\bibitem{toutanova2015observed}
K.~Toutanova and D.~Chen, ``Observed versus latent features for knowledge base
  and text inference,'' in \emph{Proceedings of the 3rd Workshop on Continuous
  Vector Space Models and their Compositionality}, 2015, pp. 57--66.

\bibitem{DBLP:journals/corr/YangYHGD14a}
B.~Yang, W.~Yih, X.~He, J.~Gao, and L.~Deng, ``Embedding entities and relations
  for learning and inference in knowledge bases,'' in \emph{Proceedings of the
  International Conference on Learning Representations (ICLR)}, 2015.

\bibitem{DBLP:conf/icml/TrouillonWRGB16}
T.~Trouillon, J.~Welbl, S.~Riedel, {\'{E}}.~Gaussier, and G.~Bouchard,
  ``Complex embeddings for simple link prediction,'' in \emph{Proceedings of
  the International Conference on Machine Learning (ICML)}, vol.~48, 2016, pp.
  2071--2080.

\bibitem{c4}
Z.~Sun, Z.~Deng, J.~Nie, and J.~Tang, ``Rotate: Knowledge graph embedding by
  relational rotation in complex space,'' in \emph{Proceedings of the
  International Conference on Learning Representations (ICLR)}, 2019.

\bibitem{DBLP:conf/acl/ChaoHWC20}
L.~Chao, J.~He, T.~Wang, and W.~Chu, ``Pairre: Knowledge graph embeddings via
  paired relation vectors,'' in \emph{Proceedings of the 59th Annual Meeting of
  the Association for Computational Linguistics and the 11th International
  Joint Conference on Natural Language Processing}, 2021, pp. 4360--4369.

\bibitem{c3}
S.~Vashishth, S.~Sanyal, V.~Nitin, and P.~P. Talukdar, ``Composition-based
  multi-relational graph convolutional networks,'' in \emph{Proceedings of the
  International Conference on Learning Representations (ICLR)}, 2020.

\bibitem{DBLP:conf/kdd/LiuGJ20}
M.~Liu, H.~Gao, and S.~Ji, ``Towards deeper graph neural networks,'' in
  \emph{Proceedings of the 26th {ACM} {SIGKDD} Conference on Knowledge
  Discovery {\&} Data Mining}, 2020, pp. 338--348.

\end{thebibliography}

\end{document}